\PassOptionsToPackage{unicode,bookmarksnumbered}{hyperref}
\PassOptionsToPackage{hyphens}{url}
\PassOptionsToPackage{dvipsnames,svgnames,x11names}{xcolor}
\documentclass[
  11pt,
  letterpaper,
]{article}
\usepackage{amsmath,amssymb}
\usepackage{lmodern}
\usepackage[sc]{mathpazo}
\usepackage{courier}
\usepackage{setspace}
\usepackage{iftex}
\ifPDFTeX
  \usepackage[T1]{fontenc}
  \usepackage[utf8]{inputenc}
  \usepackage{textcomp} % provide euro and other symbols
\else % if luatex or xetex
  \usepackage{unicode-math}
  \defaultfontfeatures{Scale=MatchLowercase}
  \defaultfontfeatures[\rmfamily]{Ligatures=TeX,Scale=1}
\fi
\IfFileExists{upquote.sty}{\usepackage{upquote}}{}
\IfFileExists{microtype.sty}{% use microtype if available
  \usepackage[]{microtype}
  \UseMicrotypeSet[protrusion]{basicmath} % disable protrusion for tt fonts
}{}
\makeatletter
\@ifundefined{KOMAClassName}{% if non-KOMA class
  \IfFileExists{parskip.sty}{%
    \usepackage{parskip}
  }{% else
    \setlength{\parindent}{0pt}
    \setlength{\parskip}{6pt plus 2pt minus 1pt}}
}{% if KOMA class
  \KOMAoptions{parskip=half}}
\makeatother
\usepackage{fancyvrb}
\usepackage{xcolor}
\IfFileExists{xurl.sty}{\usepackage{xurl}}{} % add URL line breaks if available
\IfFileExists{bookmark.sty}{\usepackage{bookmark}}{\usepackage{hyperref}}
\usepackage{pdflscape}  % adds PDF support to the landscape environment of package lscape
\definecolor{darkblue}{rgb}{0.00,0.15,0.7}
\hypersetup{
  pdftitle={Digitizing Historical Balance Sheet Data: A Practitioner's Guide},
  pdfauthor={Sergio Correia; Stephan Luck},
  pdflang={en},
  pdfkeywords={OCR, Data Extraction, Balance Sheets},
  colorlinks=true,
  linkcolor={darkblue},
  filecolor={Maroon},
  citecolor={black},
  urlcolor={darkblue},
  pdfcreator={LaTeX via pandoc}}
\VerbatimFootnotes % allow verbatim text in footnotes
\usepackage[margin=2.5cm]{geometry}
\usepackage{listings}
\newcommand{\passthrough}[1]{#1}
\usepackage{longtable,booktabs,array,threeparttable, multirow}
\RequirePackage[font=small,format=plain,labelfont=bf,textfont=it]{caption} % Nicer captions
\usepackage{calc} % for calculating minipage widths
\usepackage{etoolbox}
\makeatletter
\patchcmd\longtable{\par}{\if@noskipsec\mbox{}\fi\par}{}{}
\makeatother
\IfFileExists{footnotehyper.sty}{\usepackage{footnotehyper}}{\usepackage{footnote}}
\makesavenoteenv{longtable}

\usepackage{graphicx}
\makeatletter
\def\maxwidth{\ifdim\Gin@nat@width>\linewidth\linewidth\else\Gin@nat@width\fi}
\def\maxheight{\ifdim\Gin@nat@height>\textheight\textheight\else\Gin@nat@height\fi}
\makeatother
\setkeys{Gin}{width=\maxwidth,height=\maxheight,keepaspectratio}
\makeatletter
\def\fps@figure{htbp}
\makeatother
\providecommand{\tightlist}{%
  \setlength{\itemsep}{0pt}\setlength{\parskip}{0pt}}
\newlength{\cslhangindent}
\newlength{\csllabelwidth}
\newlength{\cslentryspacingunit} % times entry-spacing
\newenvironment{CSLReferences}[2] % #1 hanging-ident, #2 entry spacing
 {% don't indent paragraphs
  \setlength{\parindent}{0pt}
  \ifodd #1
  \let\oldpar\par
  \def\par{\hangindent=\cslhangindent\oldpar}
  \fi
  \setlength{\parskip}{#2\cslentryspacingunit}
 }%
 {}
\usepackage{calc}

\ifLuaTeX
\usepackage[bidi=basic]{babel}
\else
\usepackage[bidi=default]{babel}
\fi
\babelprovide[main,import]{english}
\def\languageshorthands#1{}

\usepackage[noabbrev]{cleveref}
\usepackage[toc,page,title]{appendix}
\usepackage{subcaption} \renewcommand{\tightlist}{} \usepackage{bm} \usepackage{econometrics} \usepackage{inconsolata}

\definecolor{codegreen}{rgb}{0,0.6,0}
\definecolor{codegray}{rgb}{0.5,0.5,0.5}
\definecolor{codepurple}{rgb}{0.58,0,0.82}
\definecolor{backcolour}{rgb}{0.99,0.99,0.98}
\usepackage{tikz}

\usetikzlibrary{patterns} \usetikzlibrary{decorations,arrows}
\usetikzlibrary{decorations.pathmorphing}
\usepgflibrary{decorations.pathreplacing}
\usetikzlibrary{arrows,positioning}
\ifLuaTeX
  \usepackage{selnolig}  % disable illegal ligatures
\fi
\usepackage[]{natbib}
\title{Digitizing Historical Balance Sheet Data: A Practitioner's
Guide\thanks{Our code is available as the
\href{https://pypi.org/project/quipucamayoc}{\passthrough{\lstinline!quipucamayoc!}}
package on the Python Package index (PyPI) as well as on Github at
\url{https://github.com/sergiocorreia/quipucamayoc/}. This paper
expresses the views of the authors and not necessarily those of the
Board of Governors of the Federal Reserve. We benefited from discussions
with Eugene White and Tom Zimmermann, as well as conference participants
at the 2021 Methodological Advances in the Extraction and Analysis of
Historical Data conference.}}
\author{Sergio Correia\footnote{Federal Reserve Board,
  \href{mailto:sergio.a.correia@frb.gov}{\nolinkurl{sergio.a.correia@frb.gov}}} \and Stephan
Luck\footnote{Federal Reserve Bank of New York,
  \href{mailto:stephan.luck@ny.frb.org}{\nolinkurl{stephan.luck@ny.frb.org}}}}
\date{\today}



\begin{document}
\maketitle

\begin{abstract}
This paper discusses how to successfully digitize large-scale historical
micro-data by augmenting optical character recognition (OCR) engines
with pre- and post-processing methods. Although OCR software has
improved dramatically in recent years due to improvements in machine
learning, off-the-shelf OCR applications still present high error rates
which limit their applications for accurate extraction of structured
information. Complementing OCR with additional methods can however
dramatically increase its success rate, making it a powerful and
cost-efficient tool for economic historians. This paper showcases these
methods and explains why they are useful. We apply them against two
large balance sheet datasets and introduce
\passthrough{\lstinline!quipucamayoc!}, a Python package containing
these methods in a unified framework.
\end{abstract}
\vspace{1.25cm}

\newcommand*{\jel}[1]{\textbf{JEL Classification:} #1}
\jel{C81, C88, N80}

\newcommand*{\keywords}[1]{\textbf{Keywords:} #1}
\pagenumbering{gobble}
\clearpage
\pagenumbering{arabic}

\setstretch{1.5}
\hypertarget{sec-introduction}{%
\section{Introduction}\label{sec-introduction}}

Optical character recognition (OCR) is a powerful tool that allows
researchers to unlock historical data. OCR quality has improved
dramatically in recent years thanks to advances in machine learning (ML)
techniques, creating opportunities to access previously unavailable
historical large-scale datasets. However, off-the-shelf OCR software
still exhibit error rates high enough to limit its application for
accurate extraction of structured information as recognition errors can
easily get compounded. For instance, applying an OCR engine with a 95
percent character accuracy rate on a table with ten six-digit numbers
will result in an output \emph{with errors} in 95.4 percent of
cases.\footnote{As a first approximation, this can be modeled as a
  binomial distribution with \(10 \times 6 = 60\) trials, each with a
  0.95 success probability.}

In this paper, we discuss how to successfully digitize large-scale
historical micro-data by augmenting OCR engines with pre- and
post-processing methods. We argue that when applying OCR, \emph{the
researcher is the practitioner}. The success of digitizing in most cases
does not depend on developing OCR engines themselves. Rather, successful
data digitization results from complementing commercial OCR software
with additional tools that can dramatically increase its success rate,
making it a powerful and cost-efficient tool for economic historians. We
illustrate why these methods are useful by applying them to two large
balance sheet datasets. Further, we introduce
\href{https://github.com/sergiocorreia/quipucamayoc}{\passthrough{\lstinline!quipucamayoc!}},
an open-source Python package containing our methods in a unified
framework.

\Cref{sec-pipeline} discusses the general trade-offs faced by
researchers when considering using OCR to digitize data. A key takeaway
is that returns to scale make OCR more practical as the underlying data
sources become larger and more standardized. In turn, the improvements
discussed in this paper, while not always cost-effective for small or
unstructured datasets, become particularly valuable for large-scale data
such as structured balance sheets. If the researcher concludes that OCR
is the most promising route to obtain her data, we thus provide a
classification and description of the general methods we recommend
using. We argue that commercial products should be integrated and
combined in a way that serves the researcher's purpose. We suggest
following a ``data extraction pipeline'' that has the following steps:
First, the original image files are pre-processed (de-warped, contrast
adjustments, etc.). Second, the commercially available OCR and layout
recognition techniques are applied. Third, the data are extracted and
validated by leveraging relationships that must hold in the data such as
accounting identities. A crucial step is a human review in which the
researcher herself validates some of the data creating a ``ground
truth.'' Such validated data allows researchers to then test and improve
the accuracy of the digitization pipeline by serving as a benchmark
against which the digitization output can be compared to construct
accuracy metrics. In turn, these metrics allow for more advanced
optimization of the parameters used throughout the digitization process.

\Cref{sec-internals} discusses each step of the data extraction pipeline
in depth. We accompany this discussion with illustrations of how these
methods were used to improve the performance of two large-scale balance
sheet digitization projects:

\begin{enumerate}
\def\labelenumi{\arabic{enumi}.}
\tightlist
\item
  The Office of the Comptroller of the Currency's (OCC) Annual Reports
  between 1867 and 1904, containing more than 100,000 national bank
  balance sheets in tabular format (\Cref{fig:example-occ}). These data
  are used, e.g., in \citet{Carlson2018}.
\item
  30,000 balance sheets of German financial and non-financial firms from
  1915 through 1933, published by \emph{Saling's Börsen-Papiere} as text
  paragraphs (\Cref{fig:example-salings}). These data are used in
  ongoing research, \citet{Brunnermeier2021}.
\end{enumerate}

% Figure generated by panflute filter "media.py"
{
\setstretch{1.0}
\begin{figure}[htpb]
  \centering
  %%%% Panel 1 %%%%
  \begin{subfigure}{0.4\textwidth}
    \centering
    \fbox{
    \includegraphics[width=0.9\linewidth]{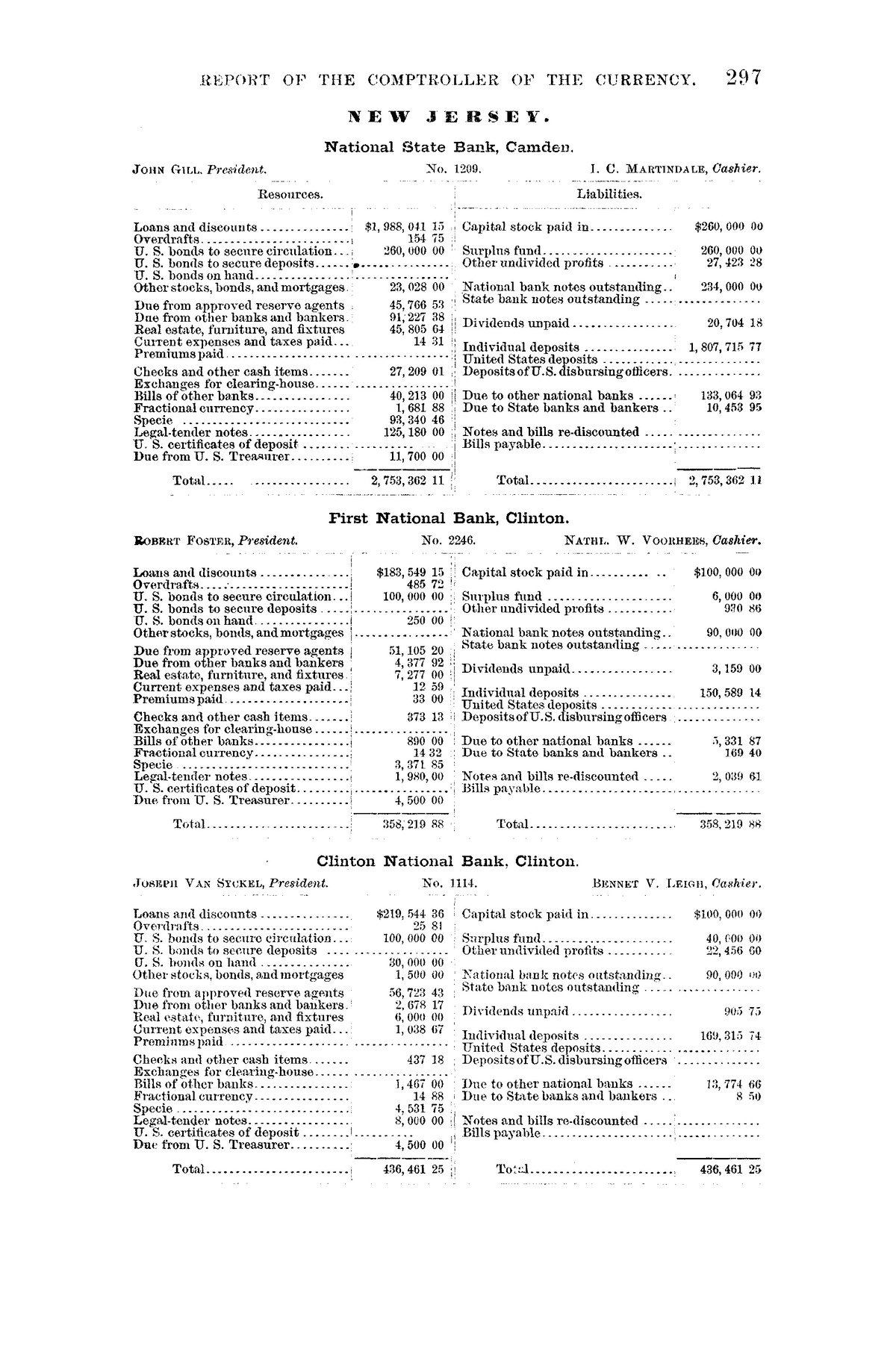}
    }
    \caption{OCC Annual Report}
    \label{fig:example-occ}
  \end{subfigure}%\hspace*{-0.1em}
  %%%% Panel 2 %%%%
  \begin{subfigure}{0.4\textwidth}
    \centering
    \includegraphics[width=0.9\linewidth]{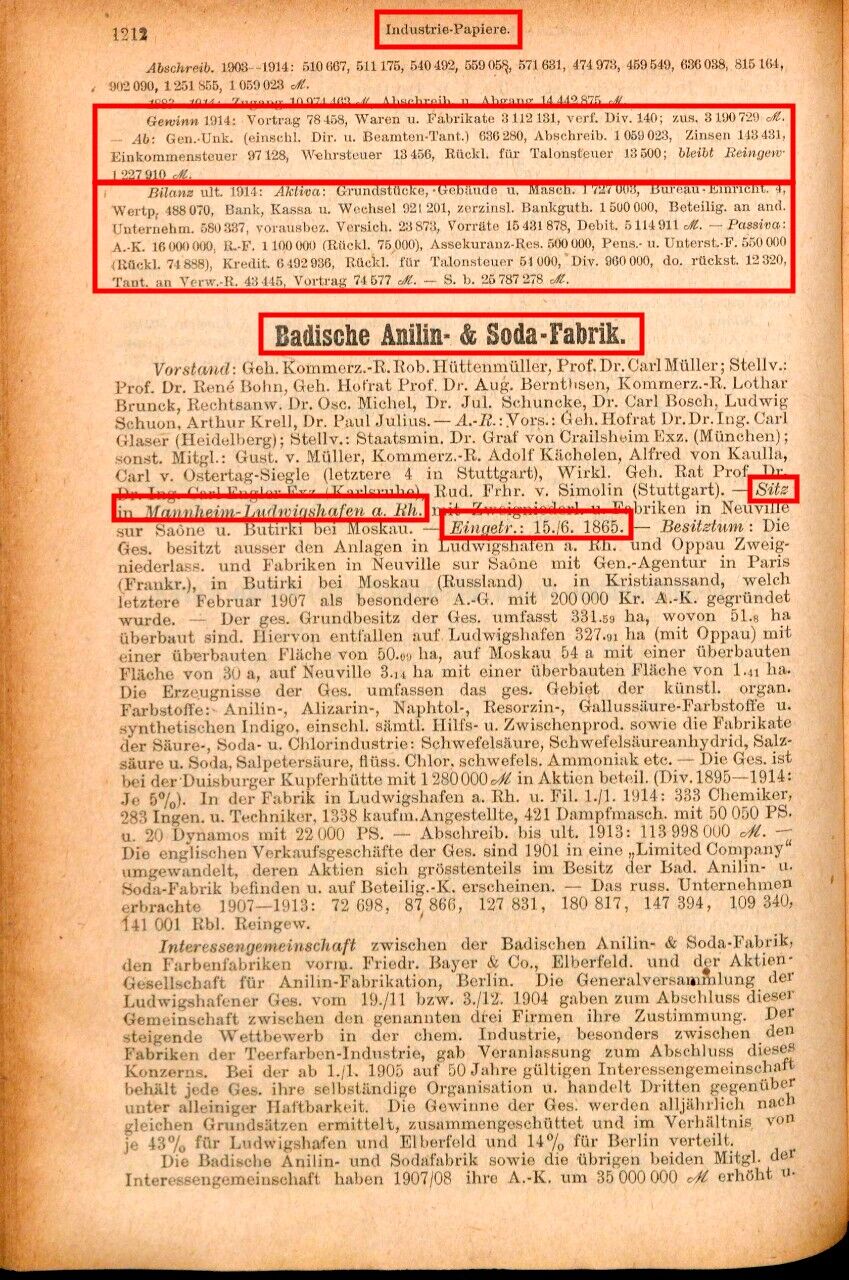}
    \caption{\emph{Saling's Börsen-Papiere}}
    \label{fig:example-salings}
  \end{subfigure}%\hspace*{-0.1em}
  \caption{\textbf{Case Studies. }These figures show examples of the two datasets showcased in this paper.
Panel (a) shows page 297 of the \emph{1882 OCC Annual Report to
Congress}. Panel (b) shows page 1212 of ``Teil 2'' of the 1915 edition
of \emph{Saling's Börsen-Papiere}, with the items to be extracted
highlighted in red.}
  \label{fig:examples}
\end{figure}
}

Although both sets of documents represent firm-level balance sheet data,
they do so in diametrically opposed layouts; the OCC dataset is in a
more standard tabular form, and the \emph{Saling's} dataset is in
free-form paragraphs. Further, these sets of documents differ in their
language, the quality of the scanned documents, the fonts used, and even
the century when they were published. This increases the likelihood that
the methods discussed here are general enough and not specific to a
certain document or type of document.

\Cref{sec-code} describes the \passthrough{\lstinline!quipucamayoc!}
Python package, showcase how to use its different components, and
illustrate how it can be used to obtain high-quality digitized data.
\Cref{sec-summary} concludes.

\hypertarget{sec-pipeline}{%
\section{Extracting balance sheet data at scale}\label{sec-pipeline}}

We start by discussing the general choices and trade-offs a researcher
faces when considering using OCR to digitize data. Depending on the size
of the historical records that will be digitized, practitioners can
generally opt for three different approaches. As shown in
\cref{figure-time-example}, these approaches differ in the initial setup
cost, as well as in the variable cost-per-page.

The least technically demanding approach is to manually input every
value into a spreadsheet. This approach does not require any coding, but
inputting each page is a slow process with roughly no returns to scale
involved: the first page takes the same amount of work as the last, and
one is not taking advantage of patterns or structures common to all the
pages to speed up the digitization process. For instance, digitizing a
single page of the OCC bank balance sheet dataset takes roughly 20
minutes per page. At this rate, digitizing the 37,000 pages comprising
the 1867-1904 OCC annual reports requires around \emph{12,300 hours} of
work.

% Figure generated by panflute filter "media.py"
{
\setstretch{1.0}
  \begin{figure}[htpb]
    \centering
    \includegraphics[width=0.6\textwidth]{"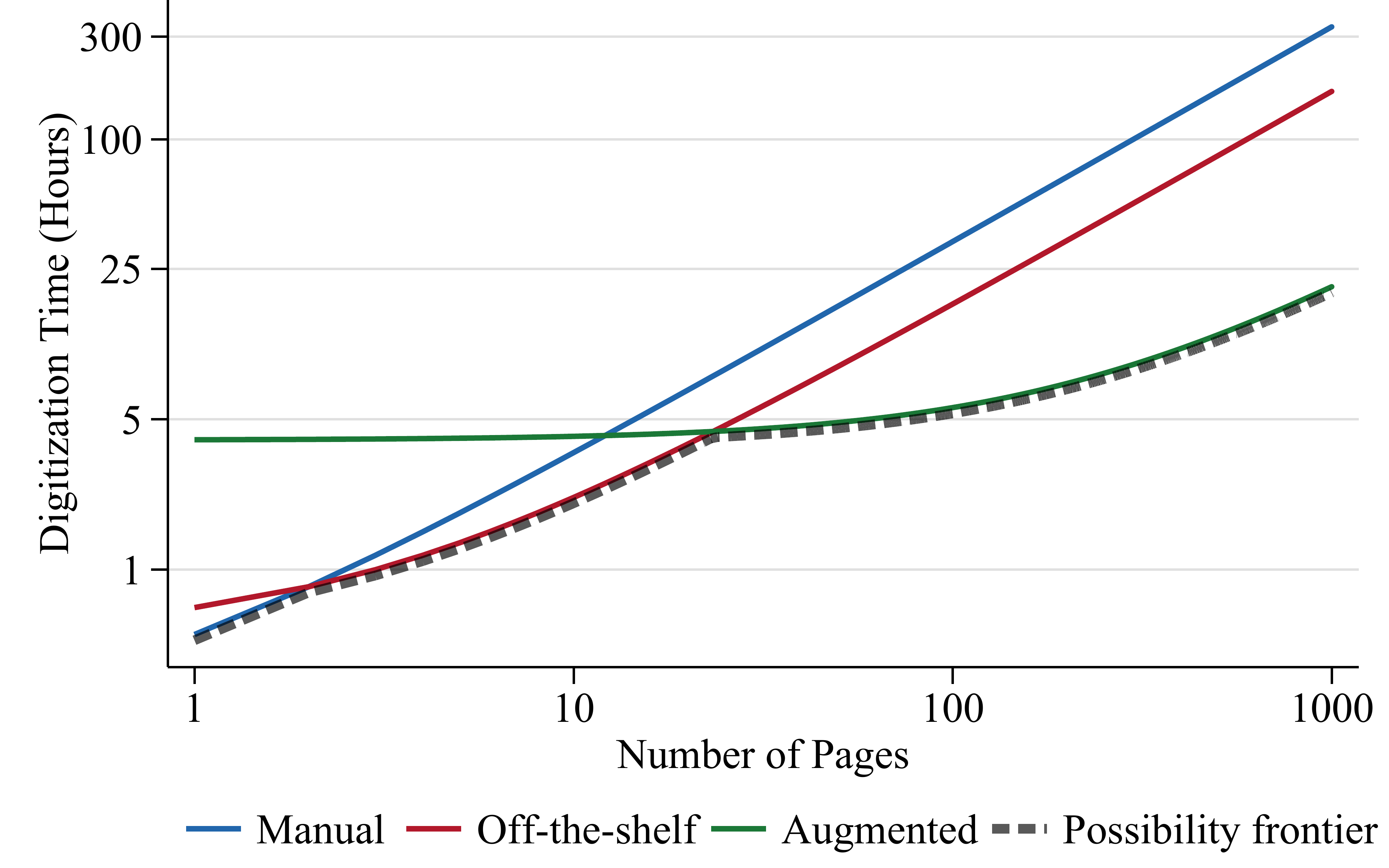"}
    \caption{\textbf{Digitization time by type of approach. }Note: axes in log-scale.}
    \label{figure-time-example}
  \end{figure}
}

A second approach, suitable for small, well-formatted
datasets---properly scanned, with standard fonts and simple
tables---consists of directly applying off-the-shelf commercial OCR
software and then manually correcting and reshaping the resulting data.
For the OCC bank balance sheet dataset, this reduced the manual review
time by roughly 60\%, to eight minutes per page. Yet such an approach is
not desirable for a large-scale digitization project; for instance
digitizing all the pages in the OCC Annual Report would require roughly
4,900 hours of manual work.

The third approach, which we recommend for large-scale datasets,
involves augmenting the OCR step with pre- and post-processing steps
that improve the quality of the resulting data and thus reduce the time
required for human review. In particular, this approach reduced the
review time of the OCC bank balance sheet to about a minute per
page---with some pages requiring extensive corrections but most
requiring minimal review. Thus, the total time required to digitize the
OCC bank balance sheet dataset becomes a much more manageable 600 hours
or roughly 15 weeks of full-time work.

Together, these three approaches form the digitization possibility
frontier, shown as the gray dashed line in \cref{figure-time-example}.
Note that because the graph is scaled in logarithms, the vast majority
of documents---with 25 pages of data or more---are best suited for the
large-scale ``augmented'' digitization approach seen in the green line.

To explain how this augmented digitization approach works, we have
classified its different elements into six broad components, illustrated
in \cref{figure-diagram}. Depending on the complexity and scale of the
input documents, researchers might want to apply only some of these
components and omit others.

% Figure generated by panflute filter "media.py"
{
\setstretch{1.0}
  \begin{figure}[htpb]
    \centering
    \resizebox{\textwidth}{!}
{
\begin{tikzpicture}

\tikzset{
    %Define standard arrow tip
    >=stealth',
    %Define style for boxes
    widebox/.style={
           rectangle,
           rounded corners,
           draw=gray, very thick,
           text width=15em,
           minimum height=2em,
           text centered},
    mediumbox/.style={
           rectangle,
           draw=black, very thick,
           text width=6em,
           minimum height=20em,
           text centered},
		smallbox/.style={
           rectangle,
           draw=black, very thick,
           text width=8em,
           minimum height=9em,
           text centered},
    circled/.style={
           circle,
           draw=black, very thick,
           text width=10em,
           text centered},
           circled2/.style={
           circle,
           draw=white, very thick,
           text width=10em,
           text centered},
}

\node[circled, fill, cyan] at (-10,0) (originaldata) {};

\node[circled2] at (-10,0) (type1) {Original Image
\begin{itemize}
\item PDF
\item PNG
\item ...
\end{itemize}
};

\node[circled,cyan, fill] at (10,0) (finaldata) {};

\node[] at (10,-1) (type1) {Data};
\draw[thick, blue, fill](10,0.5) ellipse (1.5cm and 0.75cm);
\draw[thick](10,0.5) ellipse (1.5cm and 0.75cm);
\node[] at (10,0.5) (type1) {``Ground Truth''};

\node[mediumbox] at (-5,0) (imageprocesscing) {Image Processing};

\draw[very thick,->] (originaldata) -- (imageprocesscing);

\node[smallbox] at (-1.5,2.1125) (ocrengine) {OCR Engine
\begin{itemize}
\item Engine 1
\item Engine 2
\item ...
\end{itemize}};

\draw[very thick,->] (imageprocesscing) -- (ocrengine);

\node[smallbox] at (-1.5,-2.1125) (layoutengine) {Layout Engine
\begin{itemize}
\item Engine 1
\item Engine 2
\item ...
\end{itemize}};

\draw[thick,->] (imageprocesscing) -- (layoutengine);

\node[mediumbox] at (2,0) (dataextraction) {Data Extraction and Validation};
\node[mediumbox] at (5,0) (humanreview) {Human Review};

\draw[very thick,->] (layoutengine) -- (dataextraction);

\draw[very thick,->] (ocrengine) -- (dataextraction);

\draw[very thick,->] (dataextraction) -- (humanreview);

\draw[very thick,->] (humanreview) -- (finaldata);

\node[circled] at (10,0) (finaldata) {}
edge[very thick, bend right = 60,->,gray] (imageprocesscing)
edge[very thick, bend right = 70,->,gray] (ocrengine)
edge[very thick, bend right = 70,->,gray] (dataextraction);

\node[widebox] at (5,6.5) (type1) {Parameter Tuning};

\end{tikzpicture}}
    \caption{\textbf{Data extraction pipeline. }This figure shows the six steps of the augmented digitization approach
that is the focus of this paper.}
    \label{figure-diagram}
  \end{figure}
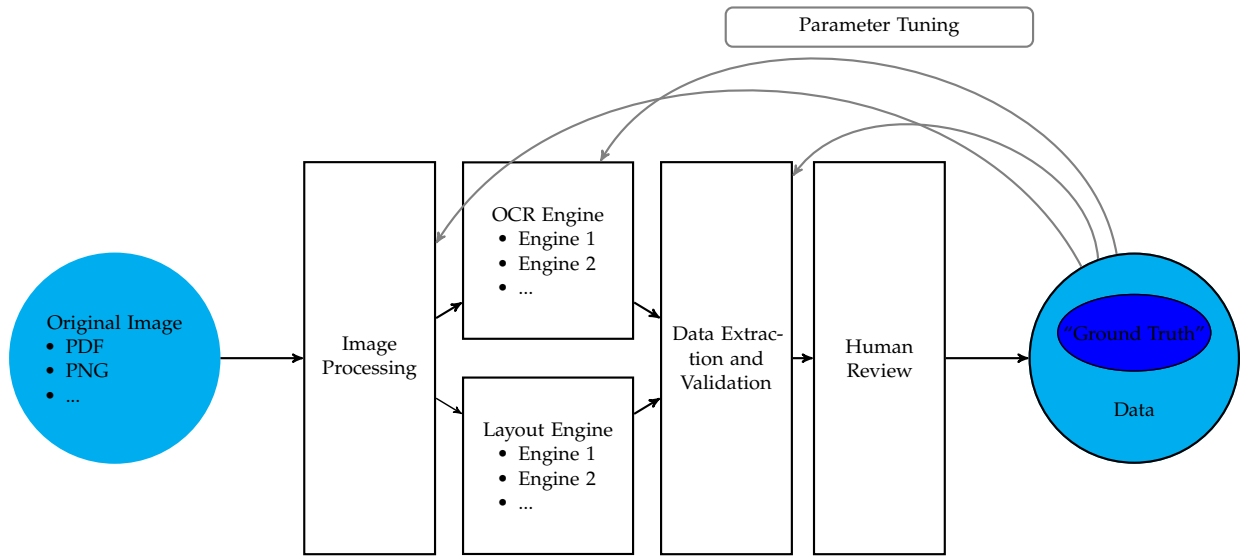
}

The starting point of this pipeline is a scanned document, either as a
PDF or as a set of images. Its endpoint is a dataset, either in tabular
form (CSV files or Excel tables) or in hierarchical form (such as a JSON
file).\footnote{Note that the resulting datasets might not just be mere
  digitization of the documents, but instead require further
  transformations to make the data useful for researchers. For instance,
  items in the \emph{Saling's} dataset are listed as free-form labels,
  so a given balance sheet might contain items such as ``Inventory of
  Portland cement at the Berlin warehouse,'' which then needs to be
  aggregated into a broader ``Inventory'' category in order to be
  comparable across firms and industries. Similarly, even the more
  standardized OCC dataset lists items as specific as ``Gold dust on
  hand,'' which we aggregate into the ``Specie'' category.} The first
step of the pipeline consists of transforming the scanned images to
improve the accuracy of the subsequent OCR steps. It consists of a
sequence of transformations known as image filters or image kernels,
where each filter has a specific purpose, such as fixing the image
alignment or increasing its contrast.

The second step consists of applying one \emph{or more} OCR engines to
the transformed images. As discussed in more detail in
\cref{subsec-ocr}, there are two reasons for using more than one engine.
First, there might not be an OCR engine uniformly better than another,
even within a single document. Rather, an engine might perform better on
some pages due to page and scanning idiosyncrasies. Thus, assuming we
can compute a page-level measure of data quality---such as the count of
numbers recognized---running multiple engines might allow us to choose
the engine best suited for a given page. Second, at an even deeper
level, we could combine multiple OCR outputs through ``ensemble
methods'' that allow us to recover a correct datum---a word or a
number---even in cases where all OCR engines outputted incorrect values.
This is feasible because most modern engines---such as Amazon's Textract
and Google's Vision AI---output not only text but its coordinates,
hierarchy---e.g.~the line, paragraph, and block where they belong---and
even confidence values of each word.

The third step of the pipeline is the layout engine, where we identify
the different layout elements of each page, such as tables, columns,
paragraphs, and headers. This step, though not often required, provides
us with two advantages. First, by knowing the section of a page where a
datum is located, we can assign it correctly within a dataset. For
instance, in the OCC example, the header of each balance sheet table
contains the name, location, and charter number of each bank. Similarly,
in the \emph{Saling's} example, the different balance sheets and profit
and loss statements are separated by paragraph breaks. The second
advantage, particularly useful for poorly scanned datasets with one or
more tables, is that by knowing where a table is located, we can crop
the image to the area corresponding to a table, and use this selection
as an input to the OCR engine, which will likely perform much better
than when inputted the entire page.

The fourth and arguably most important step for the researcher is data
extraction and validation. This step transforms the mostly unstructured
data generated by the OCR and layout recognition engines into structured
data suitable to be loaded and processed by statistical software.
Further, once the document is transformed into a structured dataset we
can evaluate the quality of its data and validate it against a set of
invariants, i.e., conditions that must always hold. For instance, the
set of labels in a balance sheet might be predefined, and its values
might be constrained to certain patterns, where ``123,456.00'' is a
valid number but ``023,456.00'' (leading zero) and ``123,4.56'' (only
one digit between comma and dot) are not.

Step five is the human validation step, which serves two purposes: The
first and most obvious is to fix incorrectly transcribed data. Depending
on the situation, researchers might want to either review all data or
data flagged as problematic if it fails sanity checks or invariants. The
second and more ancillary purpose is to construct ground truth---data
that has been reviewed and is assessed to be correct with high
confidence.

The sixth and last step is parameter tuning, which consists of adjusting
the parameters used in all the previous steps to reduce the error rate
against the ground truth available. For example, it is often not clear
what thresholds we should select when processing the images, or whether
we should apply a spell-checker. Ground truth data allows us to
construct metrics of accuracy which we can in turn use to improve the
performance of the parameters.

As we will show in the next section, OCR and data extraction and
validation are the only required steps, while the others are more
context-dependent and might be omitted depending on the quality and
scale of the documents at hand.

\hypertarget{sec-internals}{%
\section{Data extraction methods}\label{sec-internals}}

This section discusses in detail the different steps of the data
extraction pipeline outlined in \cref{sec-pipeline}. For each step, we
first discuss its rationale and the issues it addresses and then provide
an overview of some of the solutions, using as examples the two
digitization projects introduced in \cref{sec-introduction}.

\hypertarget{image-processing}{%
\subsection{Image processing}\label{image-processing}}

The main goal of this step is to undo any distortions in the digitized
images created by the scanning process. These distortions can alter the
size, shape, color, and brightness of the documents, which in turn
adversely affect the performance of the subsequent OCR step. They are
particularly problematic for historical documents, which often have
discolored pages and might suffer from artifacts such as foxing
\citep{choi2007foxing}, oxidation and fungal action stain parts of the
paper, and bleed-through \citep{GuptaGCCAGFM15}, where the ink of the
reverse page is visible in the scan of the current page. Further,
sheet-feed scanners minimize the curvature and skew of the scanned
images but are often unfeasible for historical documents as they require
removing the book spines, permanently damaging them.

\Cref{fig:examples-distortions} shows examples of the three most common
types of distortions, which we discuss in more detail below.

% Figure generated by panflute filter "media.py"
{
\setstretch{1.0}
\begin{figure}[ht!]
  \centering
  %%%% Panel 1 %%%%
  \begin{subfigure}{0.3\textwidth}
    \centering
    \includegraphics[width=0.9\linewidth]{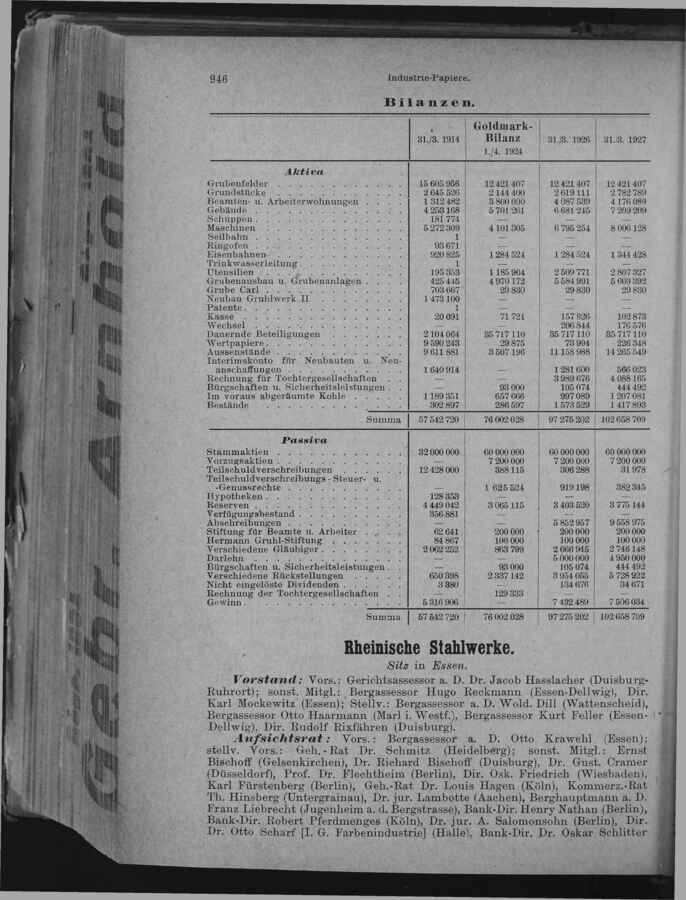}
    \caption{Distorted image size}
    \label{fig:example-distorted-size}
  \end{subfigure}%\hspace*{-0.1em}
  %%%% Panel 2 %%%%
  \begin{subfigure}{0.295\textwidth}
    \centering
    \includegraphics[width=0.9\linewidth]{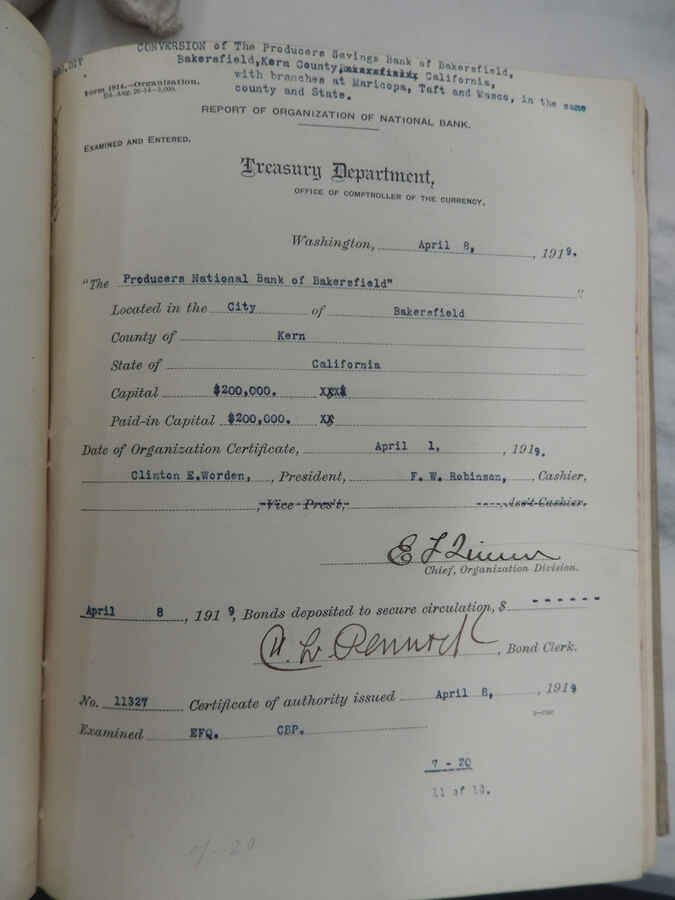}
    \caption{Distorted image shape}
    \label{fig:example-distorted-shape}
  \end{subfigure}%\hspace*{-0.1em}
  %%%% Panel 3 %%%%
  \begin{subfigure}{0.2747\textwidth}
    \centering
    \includegraphics[width=0.9\linewidth]{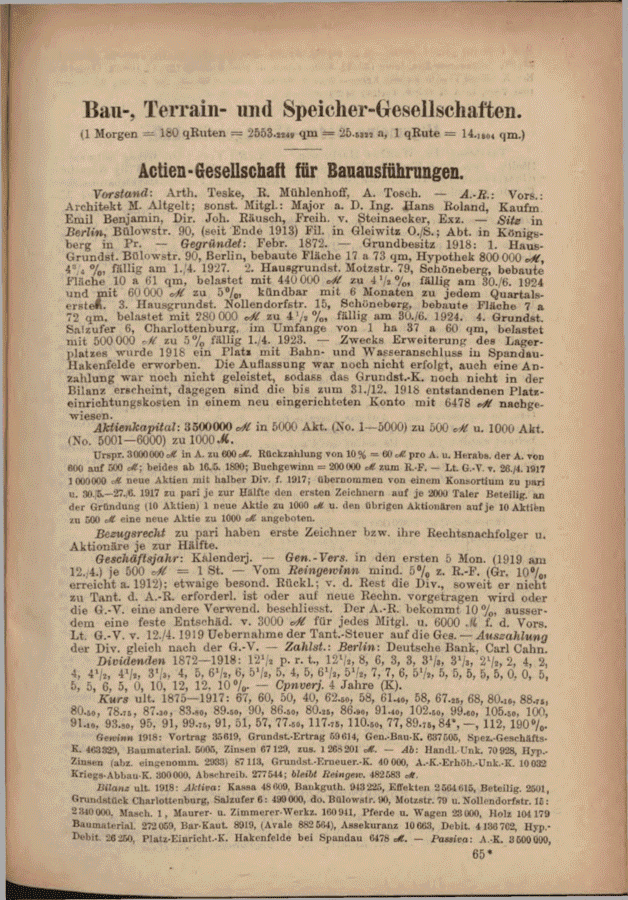}
    \caption{Distorted image color}
    \label{fig:example-distorted-color}
  \end{subfigure}%\hspace*{-0.1em}
  \caption{\textbf{Image distortions. }This figure shows examples of the three main types of image distortions
in scanned documents. Panel (a) shows size distortions in a
\emph{Saling's} page due to the fore-edge of the book and the scanner
background. Panel (b) shows geometry or shape distortions in a national
bank organization report. Panel (c) shows a \emph{Saling's} page with
yellowed background and bleed-through of the text on the reverse page.}
  \label{fig:examples-distortions}
\end{figure}
}

\hypertarget{size-distortions}{%
\subsubsection{Size distortions}\label{size-distortions}}

Due to the inability to use sheet-feed scanners, most document scanning
efforts involve either flatbed scanners or overhead cameras. These often
fail to restrict the image to the document itself, and instead produce a
scanned image that is too large and also contains part of the scanner
bed or the book edge, as shown in \cref{fig:fore1}. For the
\emph{Saling's} project, these two extraneous image elements
significantly reduced the accuracy of all OCR engines considered in this
paper.

To trim the image to correspond to the page itself, we follow a
three-part approach that proved quite robust while only requiring
minimal tuning, and is illustrated in \cref{fig:fore-edges}. This
approach exploits the fact that the page itself is in a lighter color
than the often-black scanner background and the often-gray book edge.
Then, it identifies the largest white rectangle in the image and
interprets that this rectangle corresponds to the page itself.

In more detail, the three steps are as follows:

\begin{enumerate}
\def\labelenumi{\arabic{enumi}.}
\tightlist
\item
  Binarize the image, converting it into a black-white version. This is
  implemented by converting all the pixels lighter than a certain
  threshold to white color and all others into black. Then we remove the
  noise in the image. This is implemented by sequentially applying the
  \passthrough{\lstinline!erosion!} and
  \passthrough{\lstinline!dilation!} OpenCV image filters available in
  Python \citep{Kaehler2016}. The output of this step can be seen in
  \cref{fig:fore2}.
\item
  Expand the white area to further remove noise and ensure that all
  margins of the page are white. This is implemented by using the OpenCV
  \passthrough{\lstinline!dilate!} filter. Its results are seen in
  \cref{fig:fore3}.
\item
  Identify the largest white rectangle in the image. For this, we first
  detect all rectangles with the \passthrough{\lstinline!findContours!}
  filter, which implements \citet{suzuki1985topological}. Then, we
  combine all large rectangles into a single one that in most cases
  exactly encompasses the page itself.\footnote{Note that if the
    resulting rectangle has a proportion too different from that of the
    initial document, we abort and avoid trimming the image at all. This
    is done because it is better not to crop an image than to crop too
    much of it.} The selected area is shown as a green rectangle in
  \cref{fig:fore4}, and it is this area that we use to crop the original
  document into \cref{fig:fore5}.
\end{enumerate}

% Figure generated by panflute filter "media.py"
{
\setstretch{1.0}
\begin{figure}[ht!]
  \centering
  %%%% Panel 1 %%%%
  \begin{subfigure}{0.25\textwidth}
    \centering
    \includegraphics[width=0.9\linewidth]{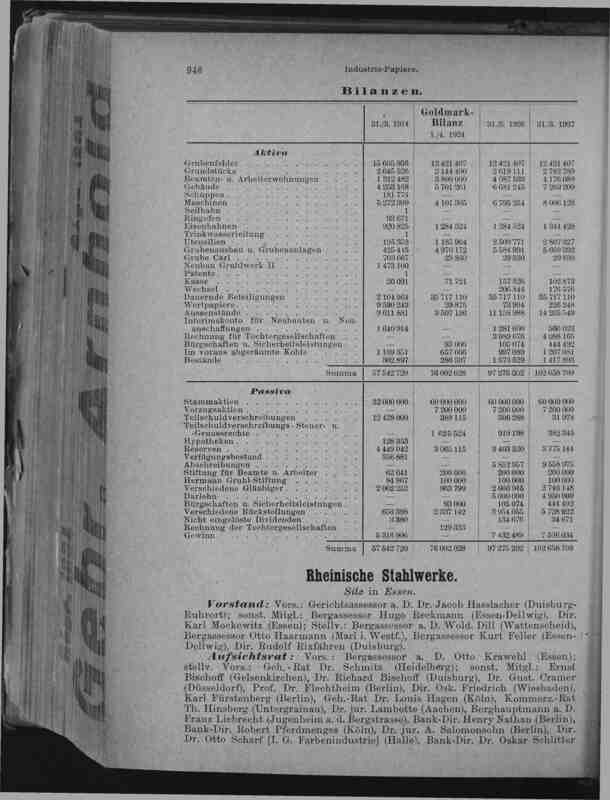}
    \caption{Input}
    \label{fig:fore1}
  \end{subfigure}%\hspace*{-0.1em}
  %%%% Panel 2 %%%%
  \begin{subfigure}{0.25\textwidth}
    \centering
    \includegraphics[width=0.9\linewidth]{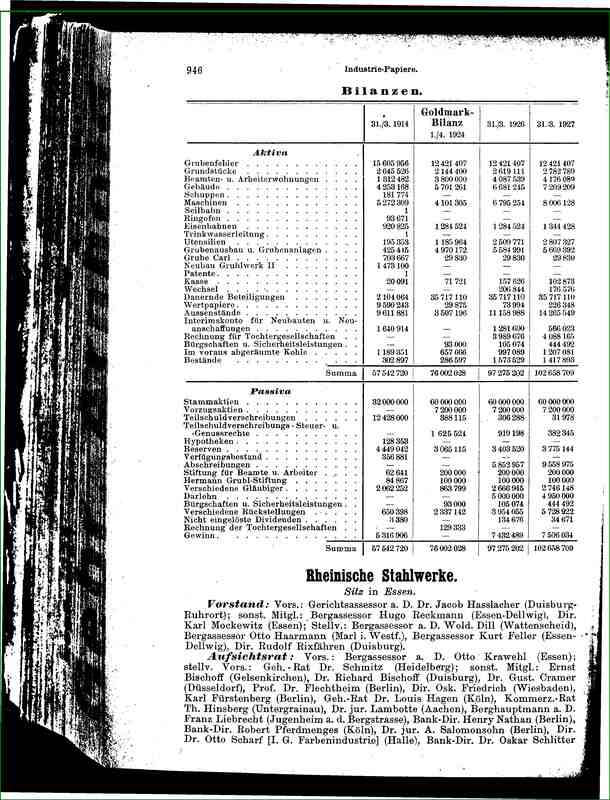}
    \caption{Binarize and remove noise}
    \label{fig:fore2}
  \end{subfigure}%\hspace*{-0.1em}
\hfill
  %%%% Panel 3 %%%%
  \begin{subfigure}{0.25\textwidth}
    \centering
    \includegraphics[width=0.9\linewidth]{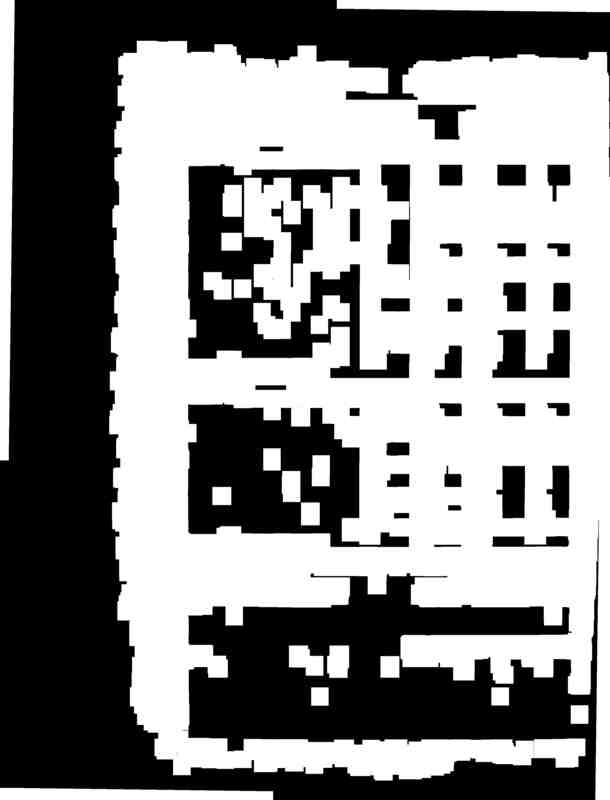}
    \caption{Dilate white area}
    \label{fig:fore3}
  \end{subfigure}%\hspace*{-0.1em}
  %%%% Panel 4 %%%%
  \begin{subfigure}{0.25\textwidth}
    \centering
    \includegraphics[width=0.9\linewidth]{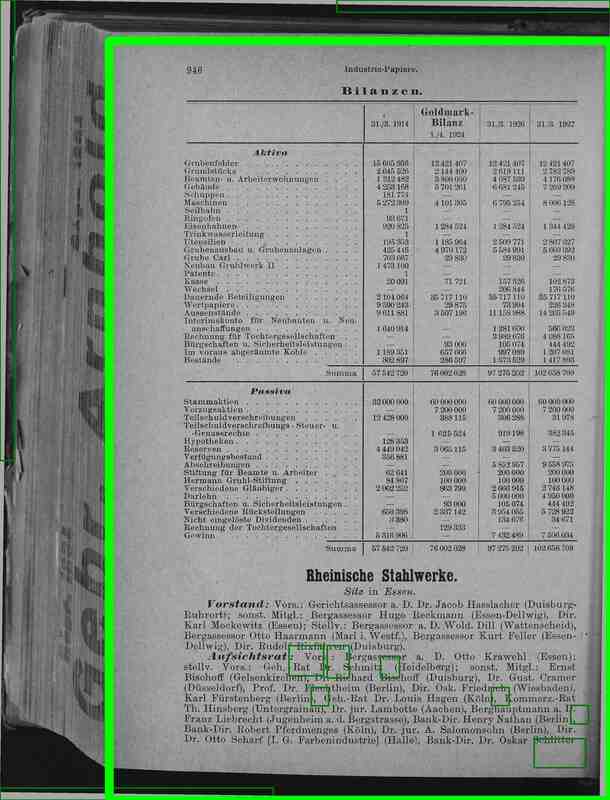}
    \caption{Identify white rectangle}
    \label{fig:fore4}
  \end{subfigure}%\hspace*{-0.1em}
  %%%% Panel 5 %%%%
  \begin{subfigure}{0.2156\textwidth}
    \centering
    \includegraphics[width=0.9\linewidth]{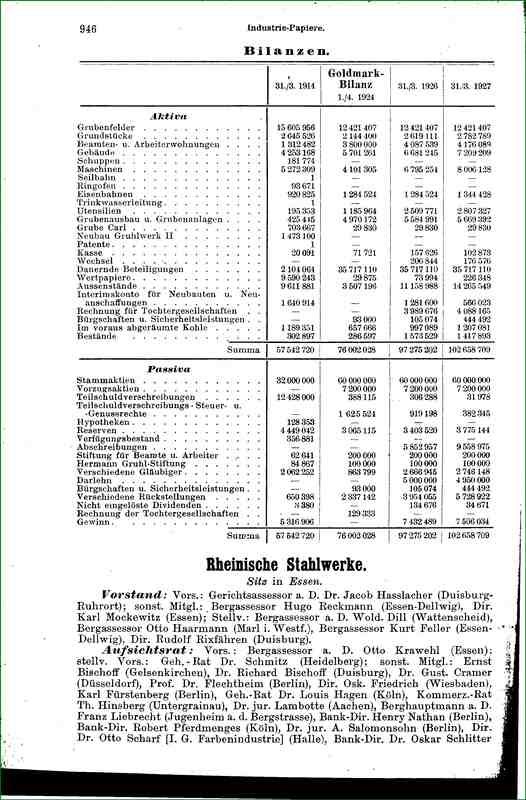}
    \caption{Trimmed output}
    \label{fig:fore5}
  \end{subfigure}%\hspace*{-0.1em}
  \caption{\textbf{Removing fore-edges on \emph{Saling's} balance sheets. }Panel (a) contains the input image. Note the poor contrast between the
text and the page background, evidenced by the narrow intensity
histogram of Panel (d). Panels (b) and (e) show the results of the
``image equalization'' method, where the intensity histogram is
stretched to span the entire black-to-white gamut. This method ignores
differences across different parts of the page, which in this case leads
to extremely dark areas close to page borders. Lastly, panels (c) and
(e) show the result of the CLAHE method, which works locally on
different areas of the page, and thus works best in cases such as where
different parts of the page had different levels of illumination, as in
this example.}
  \label{fig:fore-edges}
\end{figure}
}

\hypertarget{geometric-distortions}{%
\subsubsection{Geometric distortions}\label{geometric-distortions}}

Geometric distortions are a particularly pernicious scanning artifact
that OCR engines often fail to solve, although modern OCR engines such
as the one powering Google Books do at least partially address them with
solutions such as page dewarping \citep{GoogleBooksPatent}. These
distortions can be classified into two types.

The first, simplest case involves 2D distortions, which occur when the
scanner or camera is at an angle relative to a flat page. This is solved
by a process known as \emph{2D page dewarping and deskewing},
implemented via OpenCV's
\passthrough{\lstinline!getPerspectiveTransform!} filter, which corrects
the perspective of the camera as if the document was scanned with the
camera directly on top.\footnote{An even simpler issue is image
  rotation, which can be treated as a special case of a 2D perspective
  transform.}

The second type of distortion occurs when pages are curved while being
scanned, so not only the camera perspective has to be adjusted but the
page representation must be rectified to a flat one. Also, as discussed
in the next subsection, geometric distortions often lead to scanned
pages with non-uniform shadows, as camera lights reflect differently on
the paper depending on its curvature. Traditional approaches to
flattening curved pages involve exploiting geometric properties of
surfaces to model the curvature of a given page
\citep{tian2011rectification, Origami, dewarp-ctm}. They typically model
the entire page as a single object and assume that the curvature of the
page is smooth. Although these approaches work well for most curved
pages, they are less successful at flattening pages with folds and
creases, as the curvature is not smooth at the folds. Moreover, they
often fail to adequately remove the shadows caused by the geometric
distortions.

Recent approaches address these two limitations via machine learning
techniques. \passthrough{\lstinline!DocProj!} by \citet{PatchBasedCNN}
use a convolutional neural network to first identify distortions on
segments of a page, instead of the entire page, producing output more
robust to folds and creases. The flattened segments are then stitched,
and an additional network is used to correct the illumination of the
page. \passthrough{\lstinline!PiecewiseUnwarp!} by \citet{pwunwarp}
improve on this work by using an additional network to obtain more
robust stitching. Alternatively, \passthrough{\lstinline!DewarpNet!} by
\citet{das2019dewarpnet} exploits the fact that variations in a page
illumination are likely to be caused by geometry distortions, and thus
jointly models and corrects both types of distortions simultaneously.

\hypertarget{color-distortions}{%
\subsubsection{Color distortions}\label{color-distortions}}

Scanned historical documents often have color distortions caused by the
scanning process and age of the document. Backgrounds might be yellow
and stained due to foxing. The ink of the text on the reverse page might
be bleeding through to the current page. The illumination of the page
might not be uniform, due to poor lighting or the curvature of the
pages, particularly at the gutter shadows, which connect page edges to
the book spine.

To address this issue, the user must first decide the type of the output
image, which in most cases is either a grayscale image or a
black-and-white monochrome image. On principle, monochrome images are
preferable because the underlying document is also monochrome---black
for text and white for the background. Further, monochrome images occupy
a much smaller size and are faster to upload and process. However,
binarization, the process of converting images to black and white, often
produces poor results with low-quality scans. This is because forcing
each pixel to take binary values eliminates information that both OCR
engines and human reviewers might be able to use otherwise to better
recognize the images. Thus, the choice of grayscale or monochrome
outputs is a practical matter that depends on the characteristics of the
documents at hand. Such a phenomenon is illustrated in
\cref{fig:color-conversion}. Panel (b) shows how converting a snippet of
the \emph{Saling's} dataset to grayscale maintains the readability of
the text, while panels (c) and (d) show that although binarization
improves the clarity of most characters, some become unreadable.

% Figure generated by panflute filter "media.py"
{
\setstretch{1.0}
\begin{figure}[ht!]
  \centering
  %%%% Panel 1 %%%%
  \begin{subfigure}{0.2\textwidth}
    \centering
    \includegraphics[width=0.9\linewidth]{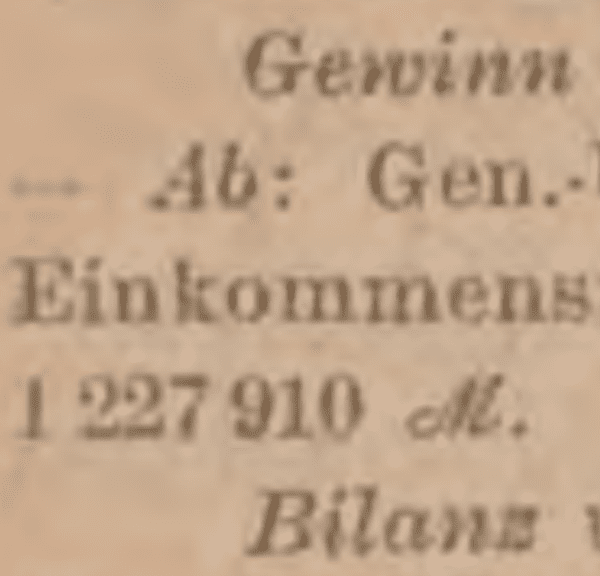}
    \caption{Input image}
    \label{fig:color-original}
  \end{subfigure}%\hspace*{-0.1em}
  %%%% Panel 2 %%%%
  \begin{subfigure}{0.2\textwidth}
    \centering
    \includegraphics[width=0.9\linewidth]{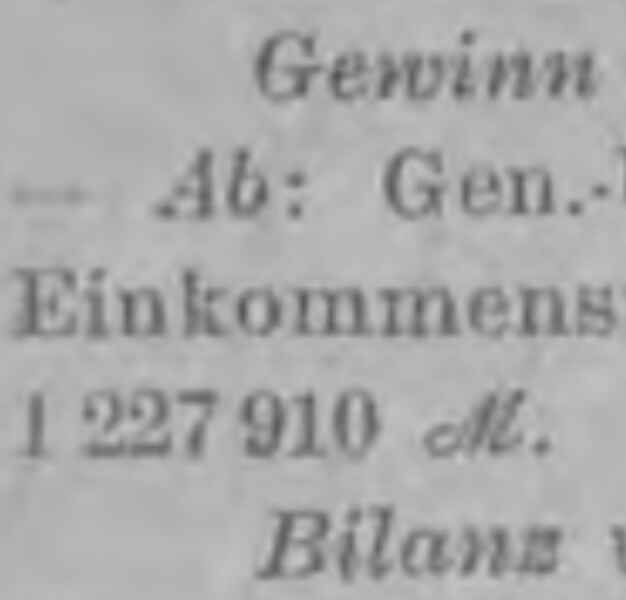}
    \caption{Grayscale}
    \label{fig:color-gray}
  \end{subfigure}%\hspace*{-0.1em}
  %%%% Panel 3 %%%%
  \begin{subfigure}{0.2\textwidth}
    \centering
    \fbox{
    \includegraphics[width=0.9\linewidth]{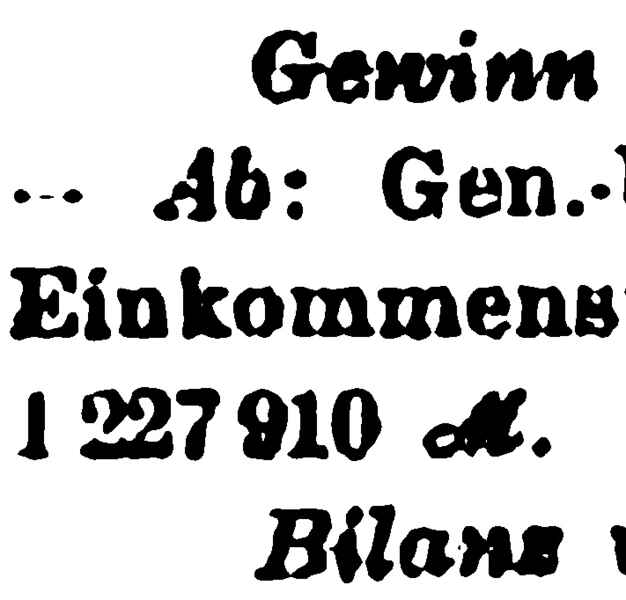}
    }
    \caption{Binarized (\(\tau=160\))}
    \label{fig:color-binary160}
  \end{subfigure}%\hspace*{-0.1em}
  %%%% Panel 4 %%%%
  \begin{subfigure}{0.2\textwidth}
    \centering
    \fbox{
    \includegraphics[width=0.9\linewidth]{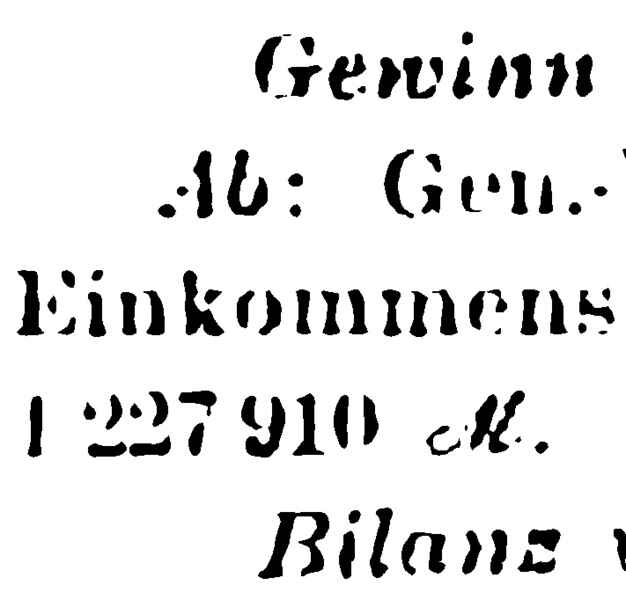}
    }
    \caption{Binarized (\(\tau=140\))}
    \label{fig:color-binary140}
  \end{subfigure}%\hspace*{-0.1em}
  \caption{\textbf{Grayscale and monochrome conversion of color images. }Panel (a) shows a zoomed-in snippet of \cref{fig:example-salings}. Panel
(b) shows its grayscale version, where all text is still easily
readable. Panel (c) shows a binarized version where all pixels with
values above 160 are converted to white and all below to black (note:
0=black and 255=white). In this panel, most letters are clearer, but
some, such as the number ``9'' or the letter ``z'' at the end of
\emph{Bilanz}, have become harder to distinguish as their font is too
thick. Lastly, panel (d) shows how reducing the binarization parameter
from 160 to 140 can produce thinner fonts, but at the cost of other
values becoming harder to distinguish.}
  \label{fig:color-conversion}
\end{figure}
}

\hypertarget{color-corrections-with-grayscale-output}{%
\subsubsection*{Color corrections with grayscale
output}\label{color-corrections-with-grayscale-output}}
\addcontentsline{toc}{subsubsection}{Color corrections with grayscale
output}

The main methods used to create color-corrected grayscale images revolve
around \emph{histogram equalization}. The starting point of these
methods is an ``intensity histogram'' characterizing all the pixels in
an image according to their intensity, from black (0) to white (255). If
an image has poor contrast, then its intensity will have a narrow
distribution which will be reflected in a histogram that is also narrow.
This can be seen in panels (a) and (d) of \cref{fig:contrast}, which
shows an image from the \emph{Saling's} dataset exhibiting poor contrast
between the page background and the text, as well as its corresponding
intensity histogram, which shows most of the figure intensity is located
at the 150-200 range.

A simple solution to the lack of contrast in this image involves
``stretching'' the intensity histogram, so it resembles more that of a
uniform distribution. This is achieved through a process known as image
equalization, showcased in \cref{fig:equalization-equalized} and
\cref{fig:equalization-equalized-hist}. There are two common problems
with this process. First, because the histogram only takes values from 0
to 255, equalizing the image will create gaps in the support, as
evidenced by looking at the large gaps between intensity values in the
100-200 range in \cref{fig:equalization-equalized-hist}. Second, the
equalization is performed uniformly across the document, so documents
that were not perfectly flat and uniformly lit when scanned will have
areas darker than others, such as the page margins, leading to
distortions in the output. This is made starkly evident in the page
margins of \cref{fig:equalization-equalized}.

To overcome these issues, we recommend instead implementing an
\emph{adaptive} histogram equalization, which creates multiple
histograms in different regions of the page. These histograms are
stretched locally, allowing these methods to better deal with
differences in lighting across the page. In particular, we recommend
using the Contrast-Limited Adaptive Histogram Equalization method, or
CLAHE \citep{CLAHE}, which adds further optimizations. For a review of
using CLAHE with historical documents, see \citet{UsageCLAHE}, which
evaluates its performance on Finnish historical newspapers.\footnote{Beyond
  equalization methods, there are other approaches aimed at solving
  particular the existence of shadows in a document. For instance,
  \citet{Bako16} implemented a novel method based on explicitly
  identifying shades and increasing their brightness.}

% Figure generated by panflute filter "media.py"
{
\setstretch{1.0}
\begin{figure}[ht!]
  \centering
  %%%% Panel 1 %%%%
  \begin{subfigure}{0.3\textwidth}
    \centering
    \includegraphics[width=0.9\linewidth]{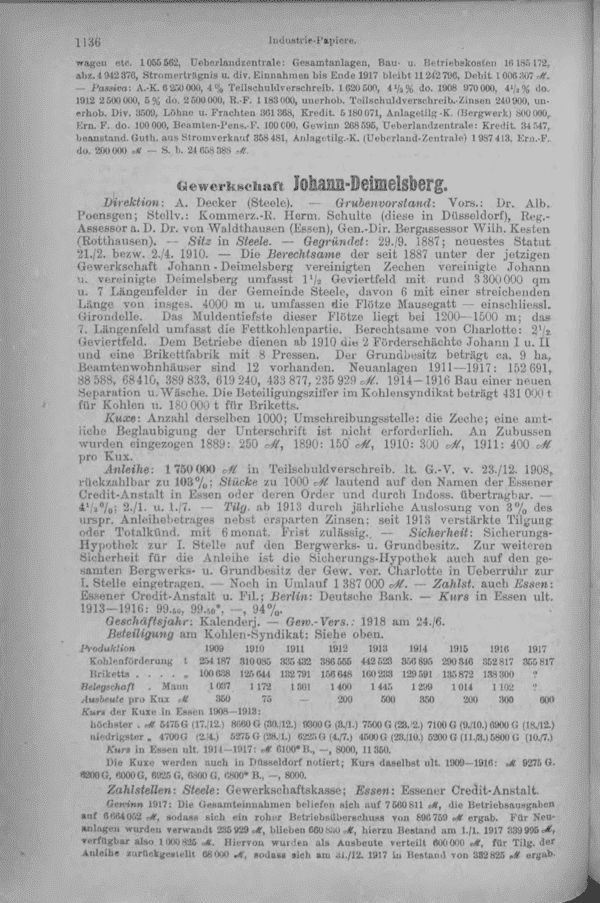}
    \caption{Input}
    \label{fig:equalization-original}
  \end{subfigure}%\hspace*{-0.1em}
  %%%% Panel 2 %%%%
  \begin{subfigure}{0.3\textwidth}
    \centering
    \includegraphics[width=0.9\linewidth]{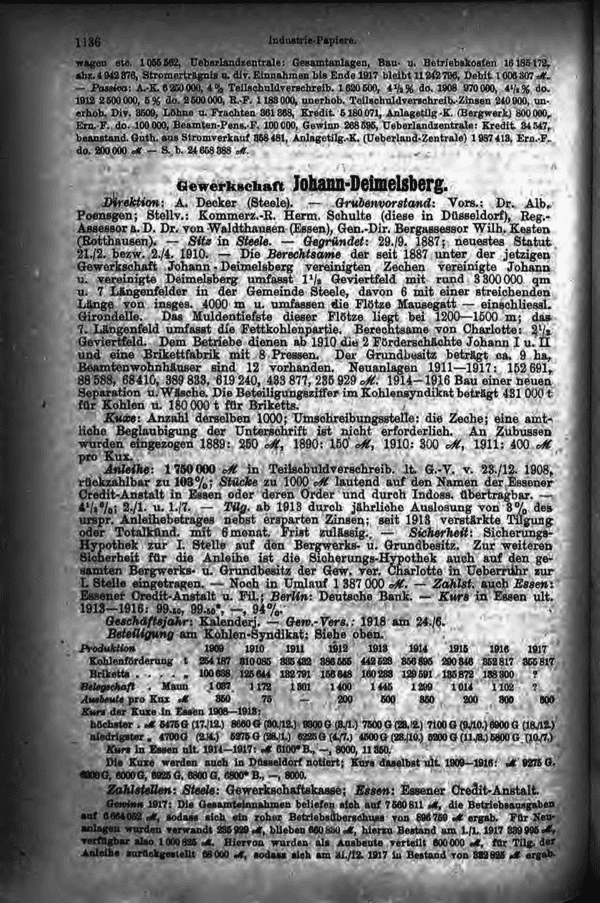}
    \caption{Equalized}
    \label{fig:equalization-equalized}
  \end{subfigure}%\hspace*{-0.1em}
  %%%% Panel 3 %%%%
  \begin{subfigure}{0.3\textwidth}
    \centering
    \includegraphics[width=0.9\linewidth]{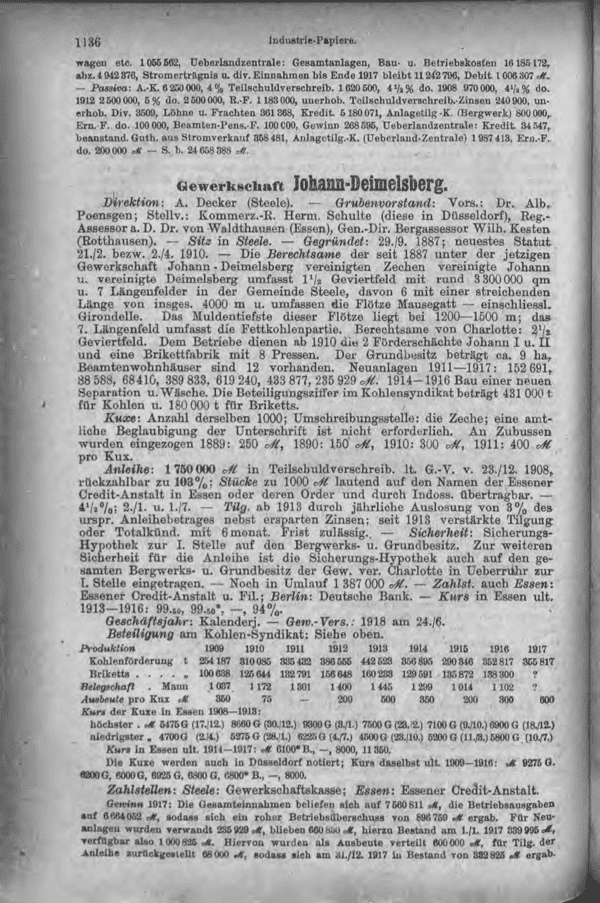}
    \caption{CLAHE}
    \label{fig:equalization-clahe}
  \end{subfigure}%\hspace*{-0.1em}
\hfill
  %%%% Panel 4 %%%%
  \begin{subfigure}{0.3\textwidth}
    \centering
    \includegraphics[width=0.9\linewidth]{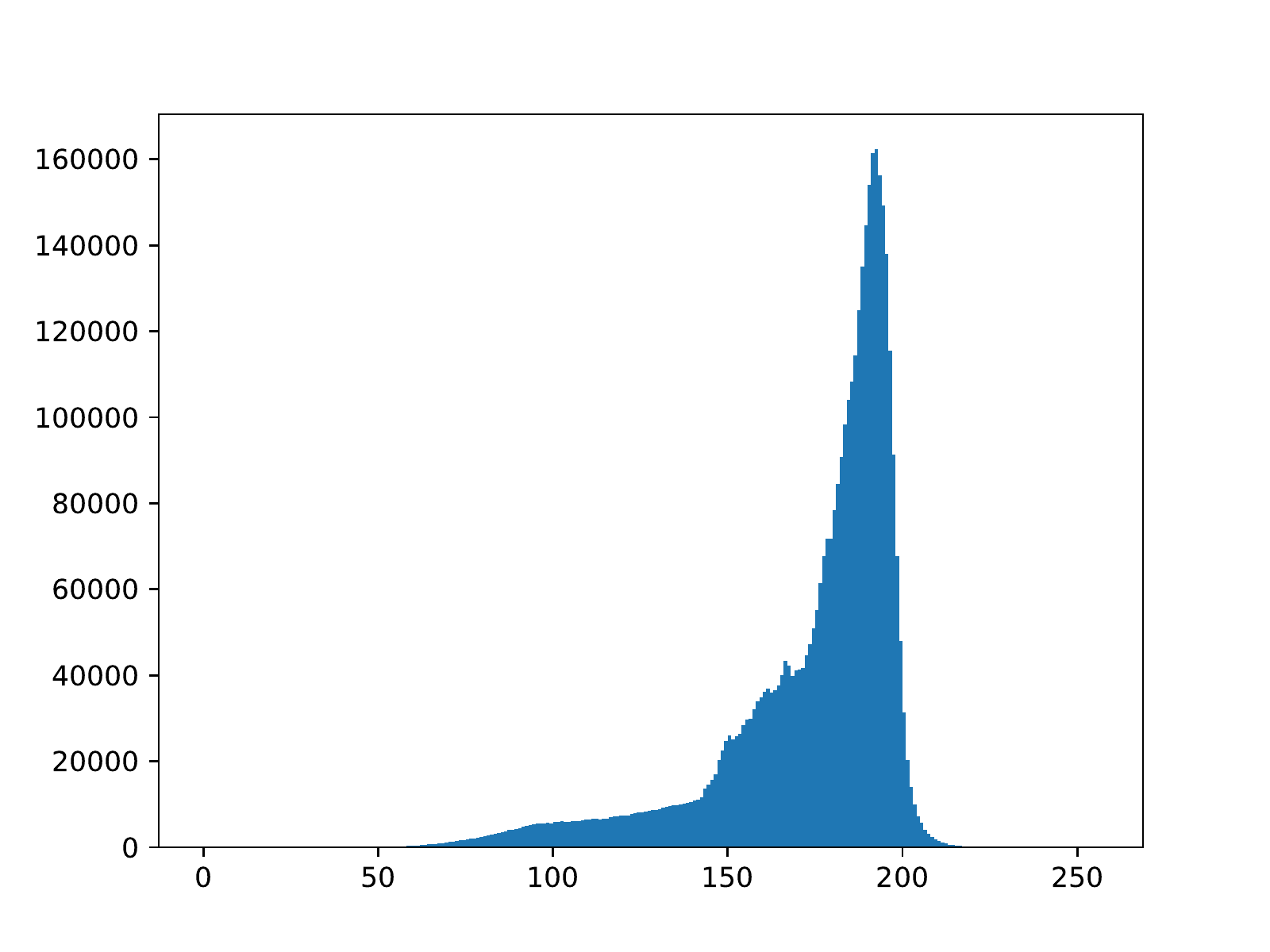}
    \caption{Histogram of input}
    \label{fig:equalization-original-hist}
  \end{subfigure}%\hspace*{-0.1em}
  %%%% Panel 5 %%%%
  \begin{subfigure}{0.3\textwidth}
    \centering
    \includegraphics[width=0.9\linewidth]{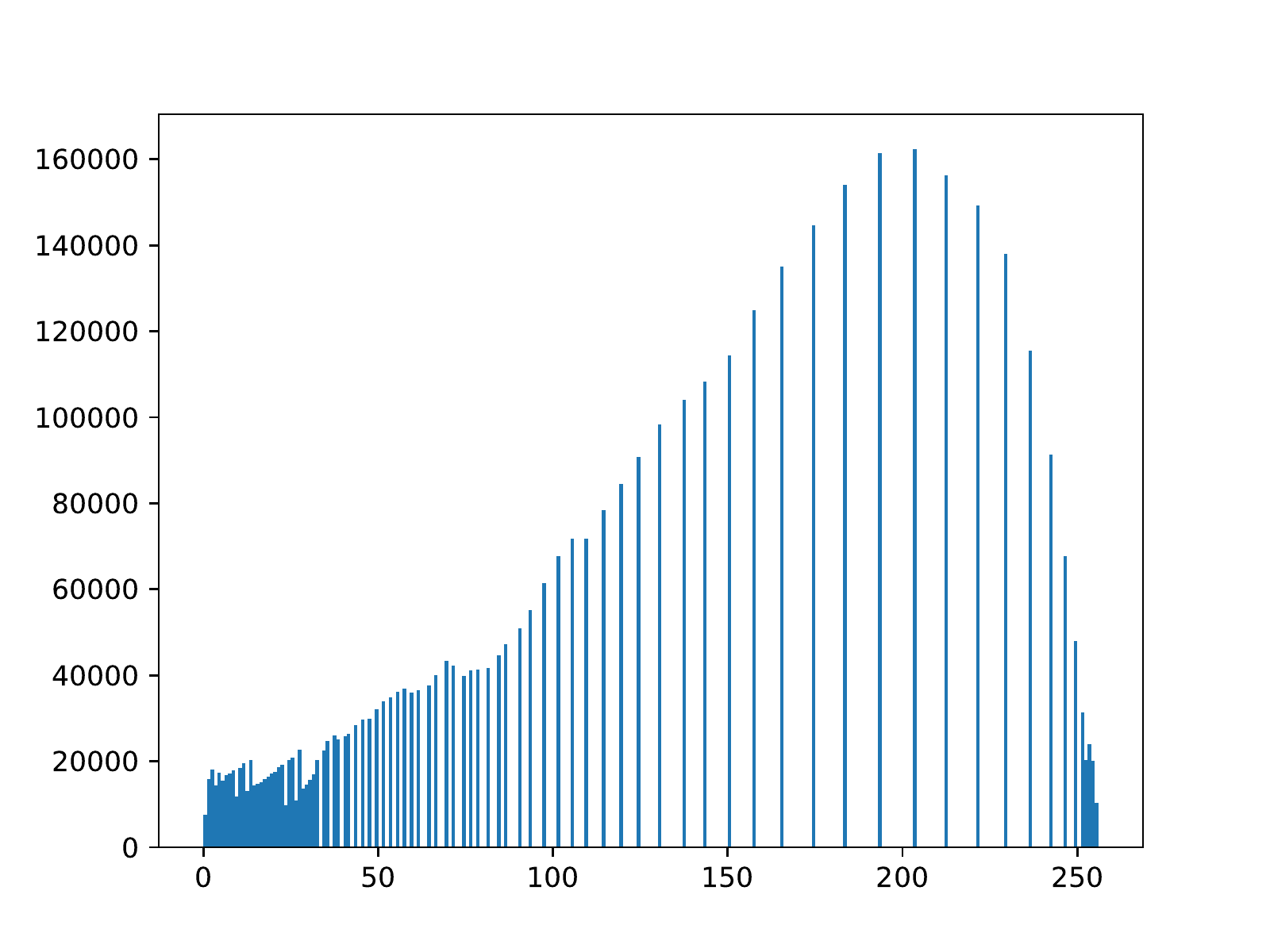}
    \caption{Histogram of equalized}
    \label{fig:equalization-equalized-hist}
  \end{subfigure}%\hspace*{-0.1em}
  %%%% Panel 6 %%%%
  \begin{subfigure}{0.3\textwidth}
    \centering
    \includegraphics[width=0.9\linewidth]{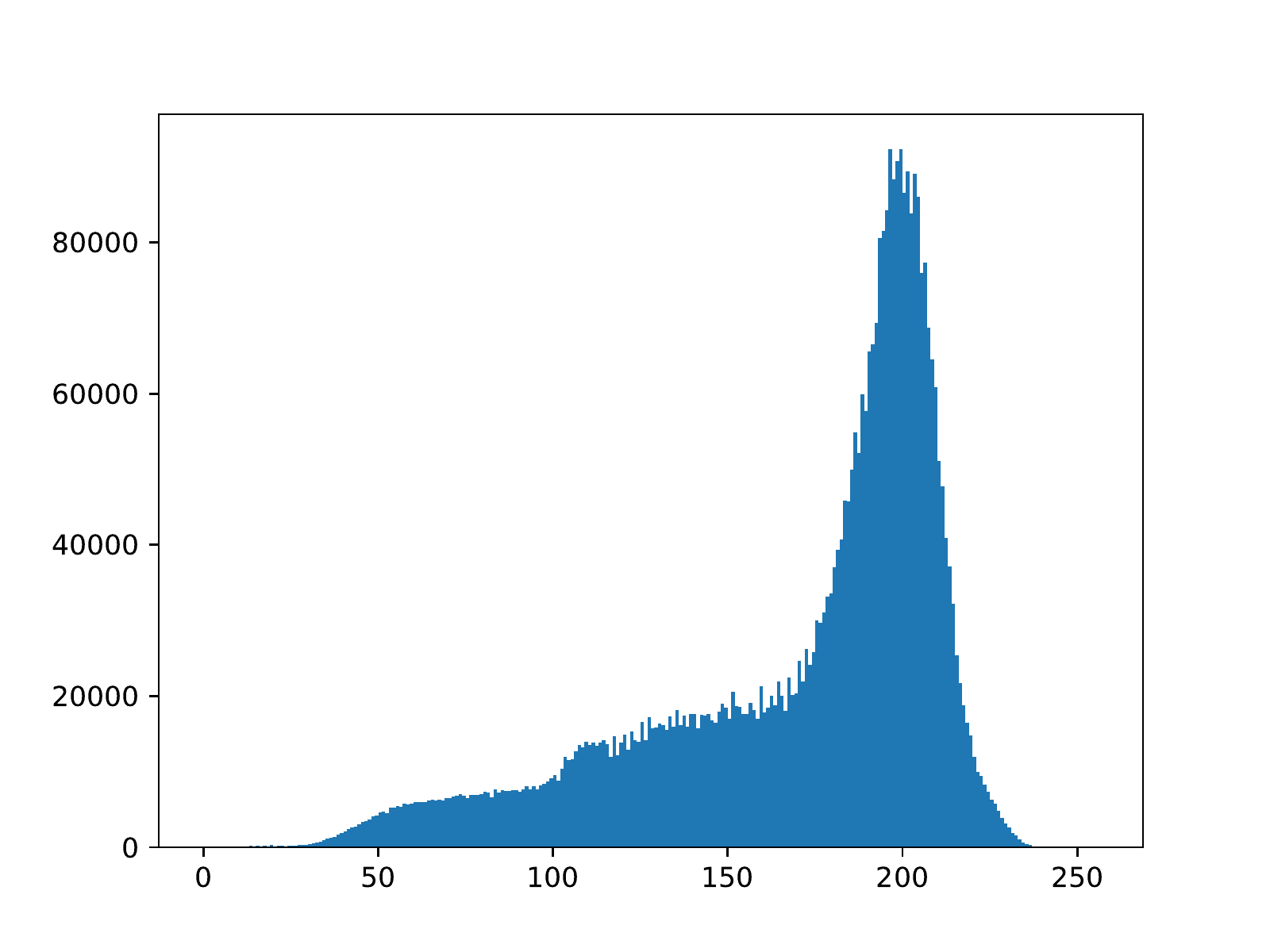}
    \caption{Histogram of CLAHE}
    \label{fig:equalization-clahe-hist}
  \end{subfigure}%\hspace*{-0.1em}
  \caption{\textbf{Improving contrast of grayscale \emph{Saling's} image. }Panel (a) contains the input image. Note the black background and the
fore-edge with the words "Gebr. Arnhold.'' Panel (b) shows the output of
the binarization and denoise step. Panel (c) (c) further reduces noise
by expanding the white space. Lastly, in panel (d) we select the largest
white rectangle in the text, corresponding to the page itself, and crop
the image to this rectangle in panel (e).}
  \label{fig:contrast}
\end{figure}
}

\hypertarget{color-corrections-with-monochrome-output}{%
\subsubsection*{Color corrections with monochrome
output}\label{color-corrections-with-monochrome-output}}
\addcontentsline{toc}{subsubsection}{Color corrections with monochrome
output}

Depending on the quality and characteristics of the scanned documents,
binarization might achieve better OCR engine performance than grayscale
conversion. Further, other methods discussed in this paper, such as line
detection, rely on a monochrome image as its input, so achieving
high-quality binarization is an essential part of the digitization
pipeline.

Nonetheless, the degradation of the paper and ink in old documents often
causes severe challenges that make binarization difficult. For instance,
page foxing and ink bleed-through can lead to white regions classified
as black after being binarized. These and other challenges are discussed
in quite extensive detail in \citet{sulaiman2019degraded}, to which we
will defer.

We explore five types of binarization methods, shown in
\cref{fig:binarization}. The first method, illustrated in
\cref{fig:binarization-threshold}, simply converts all pixels above a
predetermined threshold to white and those below to black. However, the
choice of this threshold parameter is problematic because it depends on
the characteristics of each page, and choosing the wrong threshold will
result in an illegible image. For instance, in this example, we chose as
the threshold the value 127, the midpoint between 0 and 255. This value
appears to be too low, as many font elements have been converted into
white regions.

The second method, known as Otsu's binarization, avoids this problem by
automatically choosing the threshold that minimizes the intra-class
variance of the pixel intensity. However, it performs poorly if the
document brightness is not uniform. For instance, curved documents might
be darker in the margins. That is in fact what we observe in
\cref{fig:binarization-otsu}, where we see that, although most of the
text was binarized correctly, there are nonetheless black regions in the
margins of the page.

To address brightness differences, the next set of methods are known as
local or adaptive methods, which compute thresholds at multiple areas of
the page, similarly to the adaptive histogram equalization discussed
above. In particular, \cref{fig:binarization-adaptive} implements a
mean-based adaptive binarization, \cref{fig:binarization-sauvola}
implements the method by \citet{Sauvola}, and
\cref{fig:binarization-wolf} implements the one by \citet{Wolf}. Across
these five methods, we have found the last two to be the most robust
across different types of historical documents. This finding is similar
to that of \citet{BinarizationComparison}, who compare the performance
of these and other binarization methods across documents with different
font faces, font sizes, and illumination artifacts.\footnote{For a
  broader comparison of binarization methods, see \citet{dibco2019}, who
  use a standardized benchmark to compare the performance of the 24
  methods submitted to the 2019 Competition on Document Image
  Binarization.}

% Figure generated by panflute filter "media.py"
{
\setstretch{1.0}
\begin{figure}[ht!]
  \centering
  %%%% Panel 1 %%%%
  \begin{subfigure}{0.25\textwidth}
    \centering
    \includegraphics[width=0.9\linewidth]{./compact-figures/shadow/gray.png}
    \caption{Input}
    \label{fig:binarization-original}
  \end{subfigure}%\hspace*{-0.1em}
  %%%% Panel 2 %%%%
  \begin{subfigure}{0.25\textwidth}
    \centering
    \includegraphics[width=0.9\linewidth]{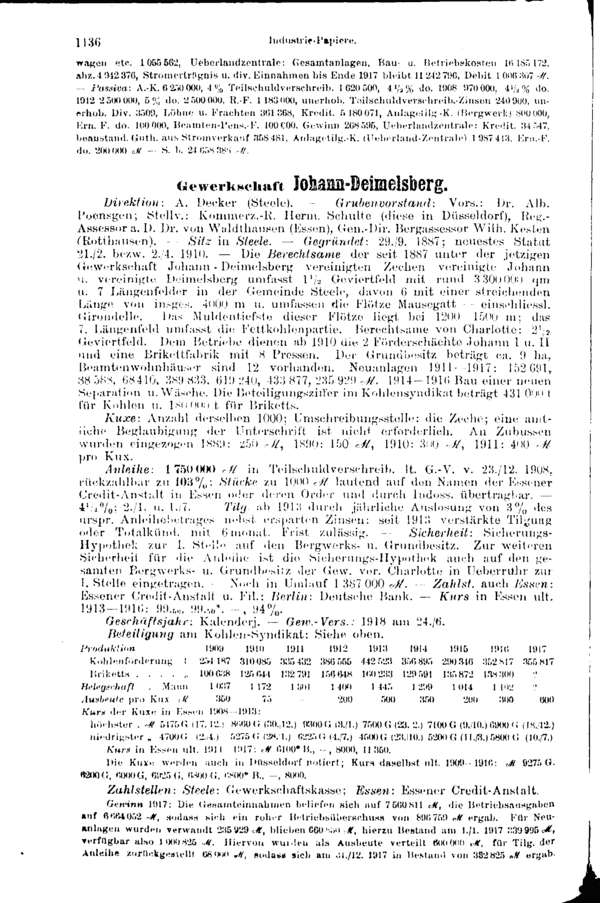}
    \caption{Threshold binarization}
    \label{fig:binarization-threshold}
  \end{subfigure}%\hspace*{-0.1em}
  %%%% Panel 3 %%%%
  \begin{subfigure}{0.25\textwidth}
    \centering
    \includegraphics[width=0.9\linewidth]{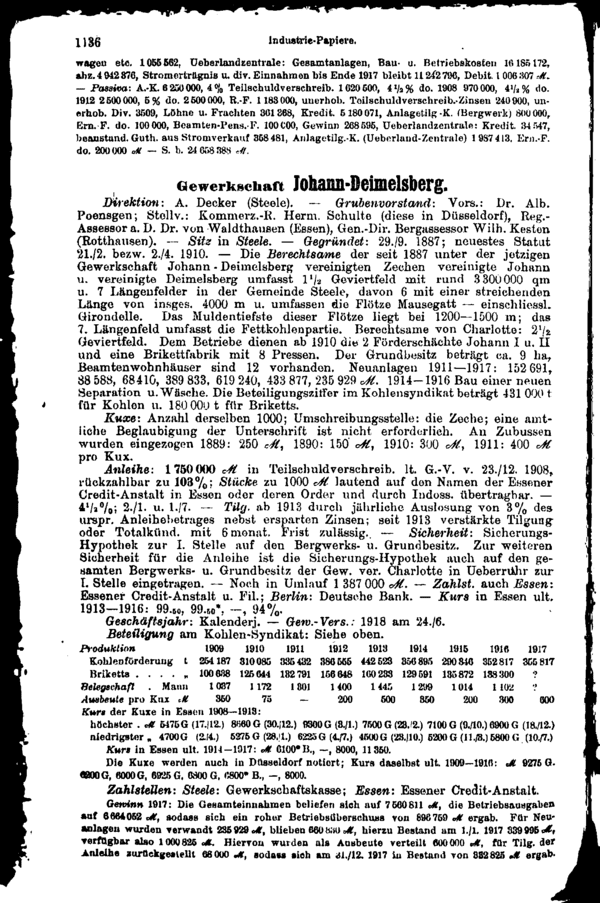}
    \caption{Otsu's binarization}
    \label{fig:binarization-otsu}
  \end{subfigure}%\hspace*{-0.1em}
\hfill
  %%%% Panel 4 %%%%
  \begin{subfigure}{0.25\textwidth}
    \centering
    \includegraphics[width=0.9\linewidth]{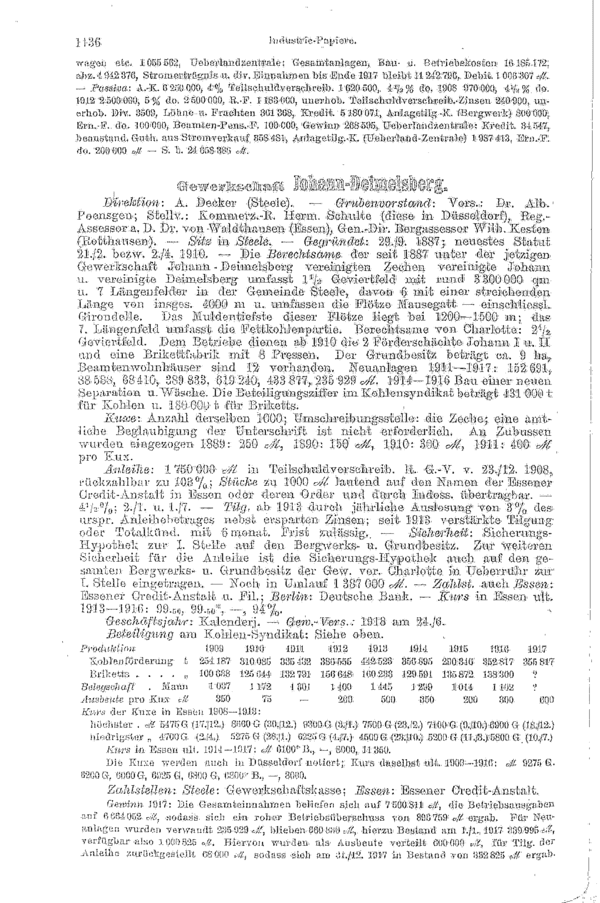}
    \caption{Adaptive mean binarization}
    \label{fig:binarization-adaptive}
  \end{subfigure}%\hspace*{-0.1em}
  %%%% Panel 5 %%%%
  \begin{subfigure}{0.25\textwidth}
    \centering
    \includegraphics[width=0.9\linewidth]{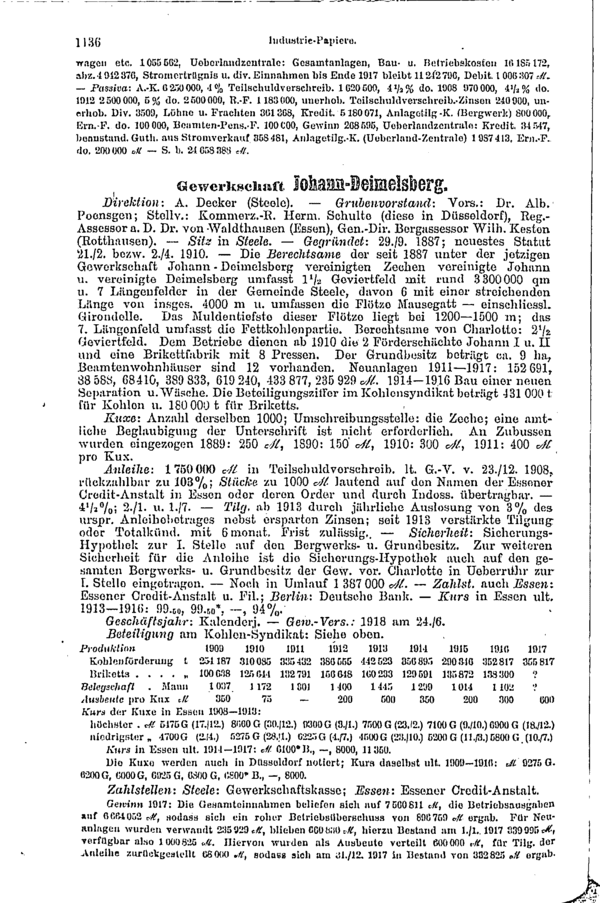}
    \caption{Sauvola binarization}
    \label{fig:binarization-sauvola}
  \end{subfigure}%\hspace*{-0.1em}
  %%%% Panel 6 %%%%
  \begin{subfigure}{0.25\textwidth}
    \centering
    \includegraphics[width=0.9\linewidth]{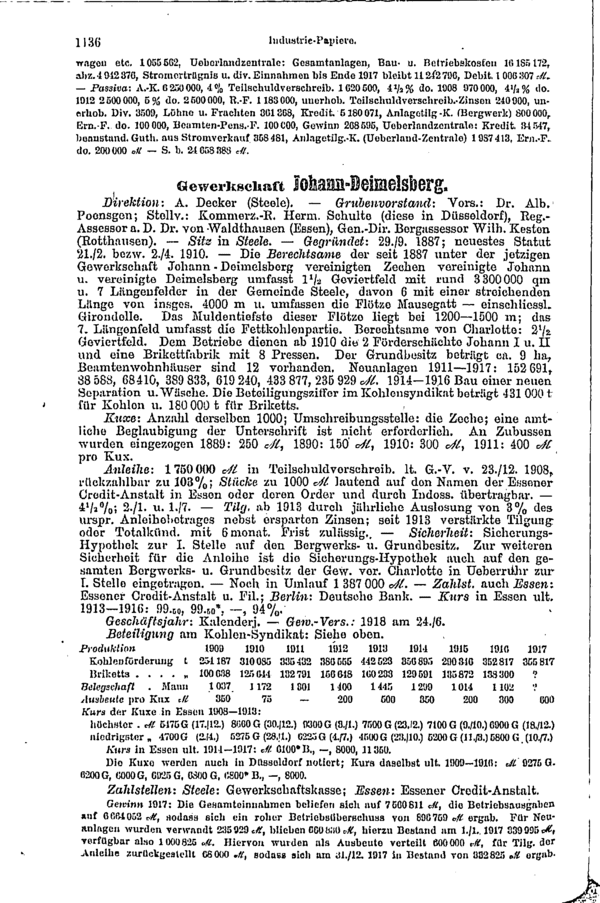}
    \caption{Wolf binarization}
    \label{fig:binarization-wolf}
  \end{subfigure}%\hspace*{-0.1em}
  \caption{\textbf{Comparison of binarization algorithms. }Panel (a) contains the input image. Panels (b) and (c) show global
binarization methods, which fail to account for brightness differences
across different page regions. Panels (d)-(f) show the performance of
three adaptive binarization methods, with methods (e) and (f) leading to
the clearest documents.}
  \label{fig:binarization}
\end{figure}
}

\hypertarget{end-to-end-methods}{%
\subsubsection{End-to-end methods}\label{end-to-end-methods}}

Recently, advances in machine learning have allowed for integrated
methods that simultaneously correct multiple types of image distortions,
often known as \emph{end-to-end} methods in the ML literature. For
instance, \citet{feng2021doctr}, implement a transformer model that
simultaneously deskews scanned documents (``geometric unwarping'') and
removes their shadows (``illumination correction'').\footnote{See this
  \href{https://github.com/fh2019ustc/DocTr}{Github repository} for a
  Python implementation.} Similarly, \citet{docentr} build an
auto-encoder transformer model that can address multiple sources of
document degradation and recover high-quality images from distorted
sources. Beyond transformers, other ML architectures are also able to
address multiple types of distortions under a single model. Thus,
\citet{degan} implement a generative adversarial network (GAN) that can
efficiently binarize documents while applying corrections for blurred
documents, watermarks, ink blobs, and ink bleed-through.

\hypertarget{subsec-ocr}{%
\subsection{Optical character recognition}\label{subsec-ocr}}

By its nature, OCR is at the core of the entire digitization process.
However, most OCR engines act as a black box and offer poor
customization, so, for practical purposes, this step is the simplest one
for the user; the only key decision is the choice of the OCR engine to
use. Although there are a large number of OCR solutions, in this section
we will focus on four widely-used OCR engines. The first three are
cloud-based commercial products from Google, Amazon, and Microsoft. The
last one, Tesseract, is an open-source engine.

As documented by \citet{Hegghammer2021}, the commercial engines vastly
outperform the open-source engines, particularly with noisy documents
likely to be found when working with historical data. Thus, our
suggestion for most users is to select a cloud-based solution, which is
likely to be both faster\footnote{Cloud providers can parallelize OCR
  tasks across many servers, while users running their own servers would
  need to either wait very large amounts of time for large-scale
  documents or manage digitization tasks across many of their own
  servers, a costly and cumbersome endeavor.} and cheaper once the
reduced time spent in the human review process is accounted for.

\Cref{table-ocr} compares four of the most widely-used OCR engines.
Strikingly, they have mostly converged in terms of their features and
the characteristics of the data they return. For instance, for each word
all offerings will return its coordinates, its confidence (likelihood of
a correct transcription), and even a hierarchy of blocks to which the
word belongs.\footnote{See also the
  \href{http://www.loc.gov/standards/alto/}{ALTO} and
  \href{http://kba.cloud/hocr-spec/1.2/}{hOCR} \citep{hOCR} standards,
  which provide XML and HTML representations of digitized material
  hierarchy and metadata.} Moreover, because they are very similar in
terms of their output, \passthrough{\lstinline!quipucamayoc!} has built
a wrapper around them so they are relatively interchangeable. This
allows users to compare the performance of alternative engines and thus
choose those which perform better on their input documents.

\begin{longtable}[]{@{}
  >{\raggedright\arraybackslash}p{(\columnwidth - 10\tabcolsep) * \real{0.1122}}
  >{\raggedright\arraybackslash}p{(\columnwidth - 10\tabcolsep) * \real{0.1735}}
  >{\centering\arraybackslash}p{(\columnwidth - 10\tabcolsep) * \real{0.1327}}
  >{\centering\arraybackslash}p{(\columnwidth - 10\tabcolsep) * \real{0.1224}}
  >{\raggedright\arraybackslash}p{(\columnwidth - 10\tabcolsep) * \real{0.3265}}
  >{\centering\arraybackslash}p{(\columnwidth - 10\tabcolsep) * \real{0.1327}}@{}}
\caption{Comparison of OCR engines\label{table-ocr}}\tabularnewline
\toprule()
\begin{minipage}[b]{\linewidth}\raggedright
Provider
\end{minipage} & \begin{minipage}[b]{\linewidth}\raggedright
Product
\end{minipage} & \begin{minipage}[b]{\linewidth}\centering
Coordinates
\end{minipage} & \begin{minipage}[b]{\linewidth}\centering
Confidence
\end{minipage} & \begin{minipage}[b]{\linewidth}\raggedright
Hierarchy
\end{minipage} & \begin{minipage}[b]{\linewidth}\centering
Lang. hints
\end{minipage} \\
\midrule()
\endfirsthead
\toprule()
\begin{minipage}[b]{\linewidth}\raggedright
Provider
\end{minipage} & \begin{minipage}[b]{\linewidth}\raggedright
Product
\end{minipage} & \begin{minipage}[b]{\linewidth}\centering
Coordinates
\end{minipage} & \begin{minipage}[b]{\linewidth}\centering
Confidence
\end{minipage} & \begin{minipage}[b]{\linewidth}\raggedright
Hierarchy
\end{minipage} & \begin{minipage}[b]{\linewidth}\centering
Lang. hints
\end{minipage} \\
\midrule()
\endhead
Google & Vision AI & Yes & Yes & Symbol, word, paragraph, block & Yes \\
Amazon & Textract & Yes & Yes & Word, line, table, cell & No \\
Microsoft & Computer Vision & Yes & Yes & Word, line & No \\
N/A & Tesseract & Yes & Yes & Word, line, paragraph, block & Yes \\
\bottomrule()
\end{longtable}

Lastly, note that beyond plain text recognition, some of the commercial
engines have begun to offer more advanced products, such as table and
form detection, as well as recognition of handwritten text. Note,
however, that these products are usually trained with modern documents,
so they often fail to perform well on historical records.

\hypertarget{ensemble-methods}{%
\subsubsection{Ensemble methods}\label{ensemble-methods}}

An implicit assumption in the paragraphs above was that users had to
choose one and only one OCR engine---the best engine---for this step.
This is not necessarily the case; it is possible to combine the output
of multiple OCR engines into an \emph{ensemble} that performs better
than any one engine by itself.

Ensemble methods go beyond looking at each page and choosing which OCR
engine performed better with the page. Instead, they go deeper into the
word level and compare the words and numbers produced by the different
OCR engines. For instance, suppose we use three engines, which identify
a given number as 123, 120, and again 123. If we interpret the engine
outputs as votes for a given candidate, then 123 would have received two
votes, 120 one vote, and thus the ensemble would have chosen 123 as its
output. Moreover, it could also be the case that the true value is 123
but the engines instead identify it as 23, 120, and 153. Here, no single
value wins the vote, so we can right-align the values and have the
engines vote at the character level, voting among blank, 1, and 1 for
the first digit, among 2, 2, and 5 for the second digit, and among 3, 0,
and 3 for the third digit. In this case, even though no single OCR
engine correctly identified the number 123, the ensemble of the three
engines still voted for the correct number.\footnote{All major engines
  provide confidence estimates on the quality of each element of a page.
  These confidences can be used to provide word- or character-specific
  weights to each vote to further improve the accuracy of the ensemble.}

There are two practical assumptions behind the use of ensemble methods
on OCR engines, without which ensembles are unlikely to provide an
advantage over single-engine approaches:

\begin{enumerate}
\def\labelenumi{\arabic{enumi}.}
\tightlist
\item
  The OCR engines are good enough to correctly identify the words, even
  though they fail to identify all their characters. If some engines
  completely fail to identify a piece of text, the accuracy of the
  method becomes limited as fewer engines would be voting on the
  results.
\item
  The errors made by the OCR engines are idiosyncratic. Otherwise, if
  all errors are systematic and common across all engines, there would
  be no advantage to using more than a single one. From experience, this
  seems to be the case, perhaps due to each product using different
  training datasets, ML models, or even different tuning parameters
  within the same model.\footnote{ Book aging, as well as the scanning
    process, provides another source of idiosyncratic errors that can be
    exploited by ensemble methods. In particular, the artifacts present
    in old books, such as discoloration or ink blobs, are often
    uncorrelated across different digital versions of the book. This is
    particularly useful with large digital libraries, such as Google
    Books and HathiTrust, which often provide multiple digital versions
    of a given book, obtained from different libraries and scanned
    independently.}
\end{enumerate}

More in-depth discussions of ensemble methods are contained in
\citet{Lund2014ensemble} and in section 3.2.1 of \citet{SurveyPostOCR}.
Further, an illustrative Stata implementation is available online
\href{https://github.com/sergiocorreia/stata-ensemble-ocr}{here}.

\hypertarget{layout-recognition}{%
\subsection{Layout recognition}\label{layout-recognition}}

The main purpose of the layout recognition step is to help researchers
assign a given word or number to specific categories or groups. For
instance, in a two-column table, identifying the location of the line
delimiting the columns would allow users to identify to which column
each word belongs. In the case of our examples, we rely mostly on two
types of layout recognition methods; one focusing on detecting lines
that delimit columns and rows and another focusing on detecting the
empty spaces that surround different sections of the text.

We illustrate the first method with the OCC dataset, where we identify
the boundaries of the three tables present on every page in order to
assign the recognized pieces of data to the correct balance sheet. As
illustrated in \cref{fig:table-lines}, this method has three steps. We
first apply a Canny Edge Detector \citep{Canny1986} to identify
boundaries between regions or objects. Then, we apply a probabilistic
Hough transform \citep{Kiryati1991, Hough1959} to the pixels highlighted
as boundary regions and identify possible lines in the text. Lastly, we
apply a custom algorithm to reduce the number of lines and avoid false
positives, and we end up with the lines identified in \cref{fig:lines1}.
With this information and the coordinates of each word produced by the
OCR engines, we are thus able to assign each word to a given column of a
table.

The second method is illustrated with the \emph{Saling's} dataset. Here,
the goal is to detect individual paragraphs, which in turn might
correspond to separate balance sheets and income tables, as well as
headers that indicate when the information for a specific firm starts.
For paragraphs, they are already pre-identified by the OCR engine, so
there is no need to implement a custom algorithm. For titles, they can
be easily identified by the coordinates of the lines (which can tell if
a line is centered), by the font size of each word, and by the empty
space above and below each line.

% Figure generated by panflute filter "media.py"
{
\setstretch{1.0}
\begin{figure}[ht!]
  \centering
  %%%% Panel 1 %%%%
  \begin{subfigure}{0.3\textwidth}
    \centering
    \fbox{
    \includegraphics[width=0.9\linewidth]{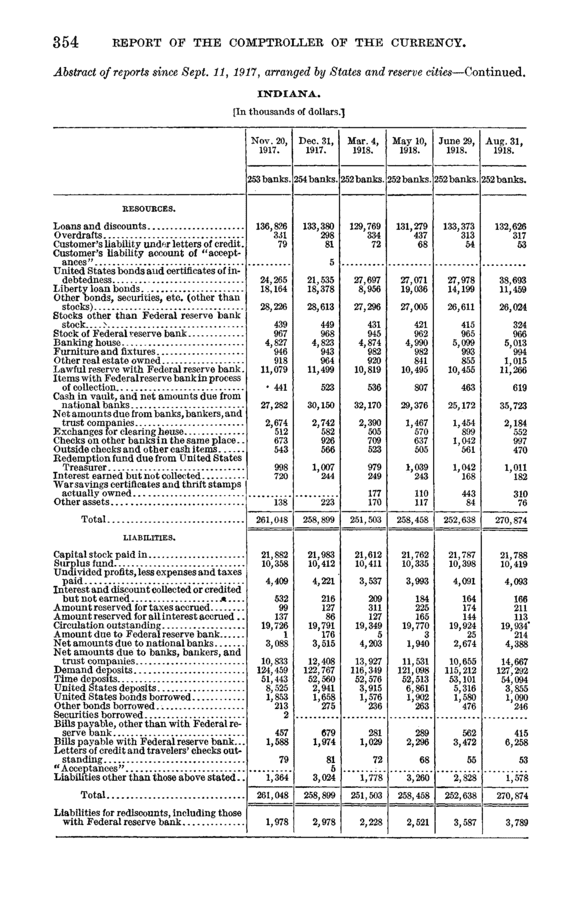}
    }
    \caption{Preprocessed image}
    \label{fig:lines1}
  \end{subfigure}%\hspace*{-0.1em}
  %%%% Panel 2 %%%%
  \begin{subfigure}{0.3\textwidth}
    \centering
    \includegraphics[width=0.9\linewidth]{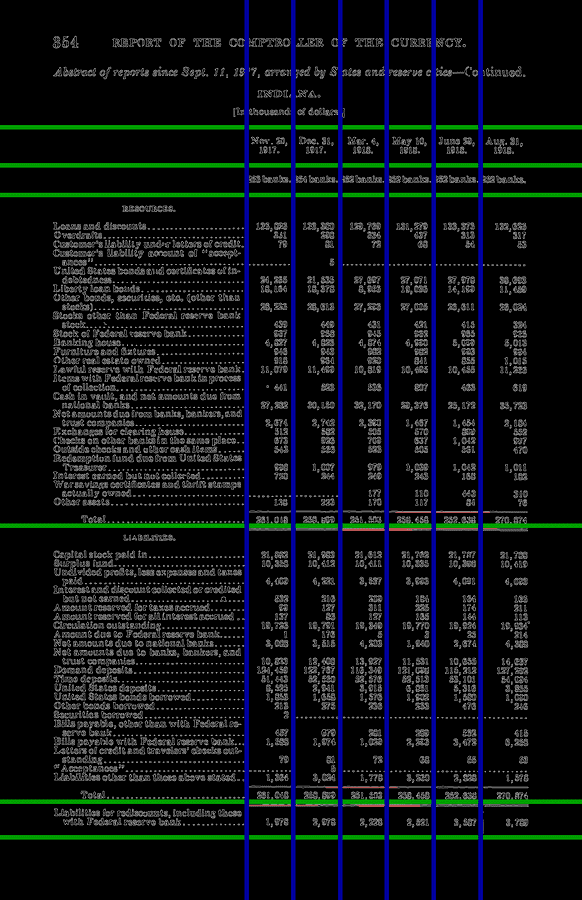}
    \caption{Image edges and detected lines}
    \label{fig:lines2}
  \end{subfigure}%\hspace*{-0.1em}
  \caption{\textbf{Detecting table delimiters on OCC balance sheets. }We combine two computer vision techniques to identify the lines that
separate the different sections of a table. First, we apply a
\href{https://en.wikipedia.org/wiki/Canny_edge_detector}{Canny edge
detector} to identify discontinuities in the brightness of the page.
Then, we apply a probabilistic
\href{https://en.wikipedia.org/wiki/Hough_transform}{Hough transform} to
identify likely lines within the text. From this set, we extract
horizontal lines (green) which detect table sections and vertical lines
(blue) which detect column delimiters.}
  \label{fig:table-lines}
\end{figure}
}

Beyond the two examples outlined above, recent advances in neural
networks have led to an almost exponential growth in the document layout
analysis literature. While these models have some disadvantages, such as
a large size due to parameters in the order of millions or billions,
they work well enough for a wide variety of layouts without requiring
ad-hoc tuning. Some of the most widely-used models include
\passthrough{\lstinline!EDD!} by \citet{EDD}, an attention-based
encoder-dual-decoder network that accurately converts table images into
tabular representations, and \passthrough{\lstinline!LGPMA!} by
\citet{LGPMA}, a deep network aimed at segmenting complex tables by
using both local information about segments of a table likely to be
table cells, and global information about the overall table layout.
Similarly, \passthrough{\lstinline!CascadeTabNet!} by
\citet{CascadeTabNet} implement a convolutional neural network model
that simultaneously detects tables in documents and identifies their
layout. However, as emphasized by \citet{LayoutParser}, note that these
models are usually pre-trained on modern document datasets, so their
performance on historical documents---which often used obsolete fonts
and design conventions such as drop caps---might not be as good. For a
comprehensive review of the performance of different machine learning
and deep neural network approaches to this topic, including discussions
on benchmarks and publicly available datasets, see
\citet{TableLitReview}.

\hypertarget{data-extraction-and-validation}{%
\subsection{Data extraction and
validation}\label{data-extraction-and-validation}}

An additional aspect of data digitization is that even if one were able
to perfectly digitize scanned data, the obtained representation most
likely would not correspond to the one required to conduct a statistical
analysis. For instance, the \emph{Saling's} balance sheets are contained
within paragraphs, so we need to identify which words correspond to
balance sheet items, which correspond to values, and which correspond to
notes or other extraneous information. A string such as \emph{``A.-K. 7
300 000, 4\% Anleihe 600 096 4 1/2\% do. 7 795 200''} would need to
first be unabbreviated (\emph{A.-K.} into \emph{Aktienkapital}),
interpreted according to the context (\emph{do.} indicating a repeat of
the previous label, i.e.~\emph{Anleihe}), and matched to a consistent
category (so both \emph{Anleihe} items would be assigned into the
\emph{Long-term debt} liability category).

More generally, the words and numbers recognized by the OCR step need to
be mapped into key-value pairs that will then be assigned to specific
cells in the final datasets. Except for very simple cases, practitioners
will inevitably have to rely on some programming to assign or reshape
the data to their needs. We believe that Python is an ideal language for
this step, as its accessible, widely known, and comes with
top-of-the-line libraries.

Given that practitioners are thus likely to be using a programming
language for the data transformation step, one might as well use this
step to reduce error rates and more generally correct mistakes caused by
the scanning and OCR processes. Otherwise, they would have to be
corrected manually, which is particularly costly and slow. Below we
describe two types of improvements that we have used in the OCC and
\emph{Saling's} digitization efforts, and which we believe might be
generally useful for practitioners.\footnote{A novel field, known as
  visually-rich Document Understanding (VrDU), combines the layout
  recognition and data extraction steps, directly producing information
  from scanned structured documents, such as forms or receipts. For
  instance, \passthrough{\lstinline!LayoutLMv2!} by \citet{LayoutLMv2}
  is a multimodal transformer model that combines textual, layout, and
  image information to extract data from scanned documents, such as
  entities or names. Note however that these models might not be as
  practical for historical data extraction; for instance, the list of
  entities used to train the models might be vastly different than those
  in historical records.}

\hypertarget{correction-of-words-and-numbers}{%
\subsubsection{Correction of words and
numbers}\label{correction-of-words-and-numbers}}

There are often restrictions on the possible values that each key or
value can have. For instance, numbers representing dollar values cannot
contain letters. Thus, if the letter ``O'' was found between two digits
(``1O9''), it would be prudent to replace it with the number zero.
Similar patterns apply to other letters, so ``1GB'' could be replaced
with ``168.'' Note, however, that there is a cost to these replacements
in that they might introduce false positives, so users should avoid
being too aggressive with them and always try to benchmark how the
addition of each replacement affects the quality of the final dataset.

Labels often only take a small set of possible values. For instance, in
most years, the OCC balance sheets contained only a few dozen possible
asset and liability categories. Similarly, the city where each bank was
located---written at the top of each table---could be validated against
the list of all existing and ghost towns in the US maintained by the
U.S. Geological Survey. As a last resort, one could validate a free-form
word against the set of valid words in a given language---its
dictionary---to assert whether the word is valid or not.

Once a word has been diagnosed as invalid, it can be fixed by applying a
spellchecker to it. For instance, Peter Norvig's famous spellchecker
\citep[see][]{Kemighan1990, Jurafsky2009} can be easily implemented in a
few lines of code as long as there is a dictionary with the list of all
valid values.

The improvements we have discussed are mostly specific to our specific
datasets, but can be expanded into more general approaches.
\citet{PostCorrection} propose an automatic post-OCR correction process
that works in two stages. First, independent modules propose corrections
to the OCR output. These corrections can be at the word level, such as
spell checkers and word splitters, at the sentence or word pair level,
such as word mergers, and even at the text level, where probabilistic
spellcheckers suggest corrections based on the internal frequency of
words within the text. In the second stage, a decision model chooses the
most likely correction among the proposed candidates.

Some approaches combine this step with ensemble methods discussed
earlier, particularly for OCR engines also produce confidence estimates
of their result. An example of such an approach is \citet{Classical},
who implement an ensemble through a technique called ``progressive
multiple alignment,'' where OCR output from multiple engines are
aligned, and then the most likely character across the aligned datasets
is chosen. This is then followed by a spellchecker based on a naive
Bayes classifier, or with a regular expression in case the spellchecker
fails to find a match.

For a broader discussion of the post-OCR processing problem, see
\citet{SurveyPostOCR}, who provide an in-depth discussion of the
post-OCR processing problem, illustrate typical pipelines, and compare
several existing approaches, ranging from manual approaches such as
ReCAPTCHA to state-of-the-art neural network language models. For a
survey tailored towards historical texts, see \citet{Bollmann} does, who
surveys the particular difficulties of normalizing historical documents,
where for instance word spellings change as time passes (e.g.~``their''
was once ``theyr'', ``thar'', or ``þer''). The paper then compares
solutions involving rule-based methods, statistical models, and neural
network approaches. Interestingly, and in contrast to other steps where
neural networks have proven substantially more effective, the author
finds that statistical models and even rule-based methods might have
similar performance, which makes them preferred given the high computing
costs required by deep neural networks, particularly at the training
stage.

\hypertarget{correction-of-document-hierarchy}{%
\subsubsection{Correction of document
hierarchy}\label{correction-of-document-hierarchy}}

As discussed in the previous subsection, most OCR engines provide
incredibly useful information on the hierarchy to which each word
belongs---its line and paragraph. However, in the same way as words
might be misrecognized, this hierarchy can also be detected incorrectly.
This is illustrated in \cref{fig:table-rows}. There, \cref{fig:rows1}
shows all the words and paragraphs identified by Google's Vision AI.
Because the page represents a table, the paragraphs are not recognized
correctly although the words are.

To solve this, the first step, shown in \cref{fig:rows2}, involves
grouping words together into lines, which will represent the balance
sheet labels. For this, we simply pair up words in the same column as
long as they have enough overlap in the y-axis. However, several labels
are long enough to overflow the column width and thus occupy multiple
labels. To solve this, we exploit the fact that the second line in each
label is always indented, which allows us to arrive at the corrected
output in \cref{fig:rows3}.

% Figure generated by panflute filter "media.py"
{
\setstretch{1.0}
\begin{figure}[ht!]
  \centering
  %%%% Panel 1 %%%%
  \begin{subfigure}{0.3\textwidth}
    \centering
    \includegraphics[width=0.9\linewidth]{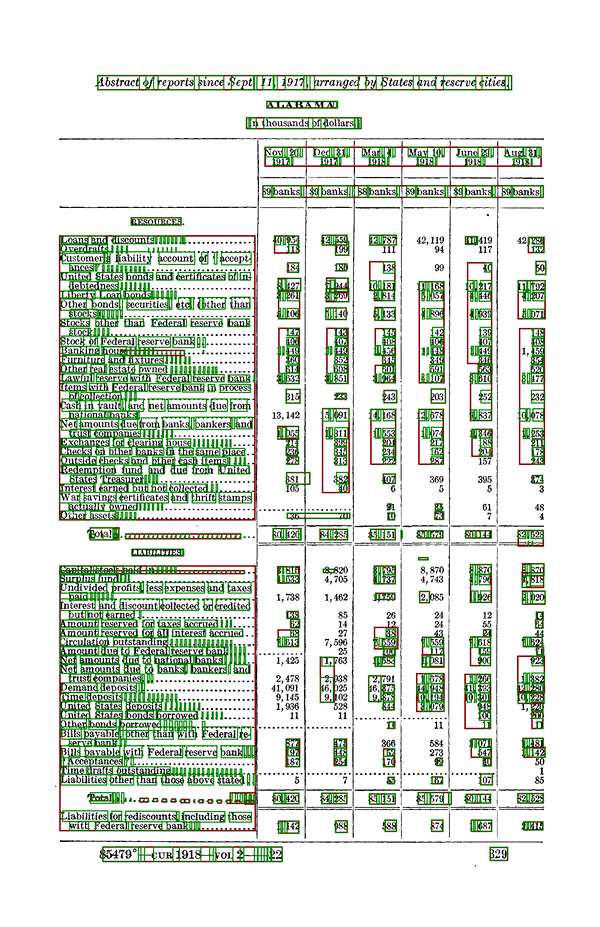}
    \caption{OCR-recognized ``words''}
    \label{fig:rows1}
  \end{subfigure}%\hspace*{-0.1em}
  %%%% Panel 2 %%%%
  \begin{subfigure}{0.3\textwidth}
    \centering
    \includegraphics[width=0.9\linewidth]{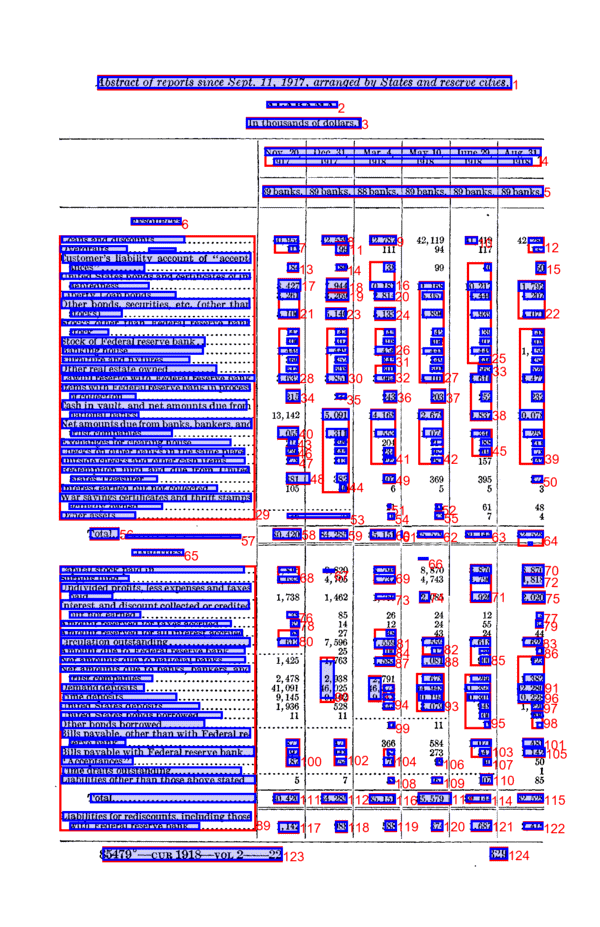}
    \caption{Lines identified by \texttt{quipucamayoc}}
    \label{fig:rows2}
  \end{subfigure}%\hspace*{-0.1em}
  %%%% Panel 3 %%%%
  \begin{subfigure}{0.3\textwidth}
    \centering
    \includegraphics[width=0.9\linewidth]{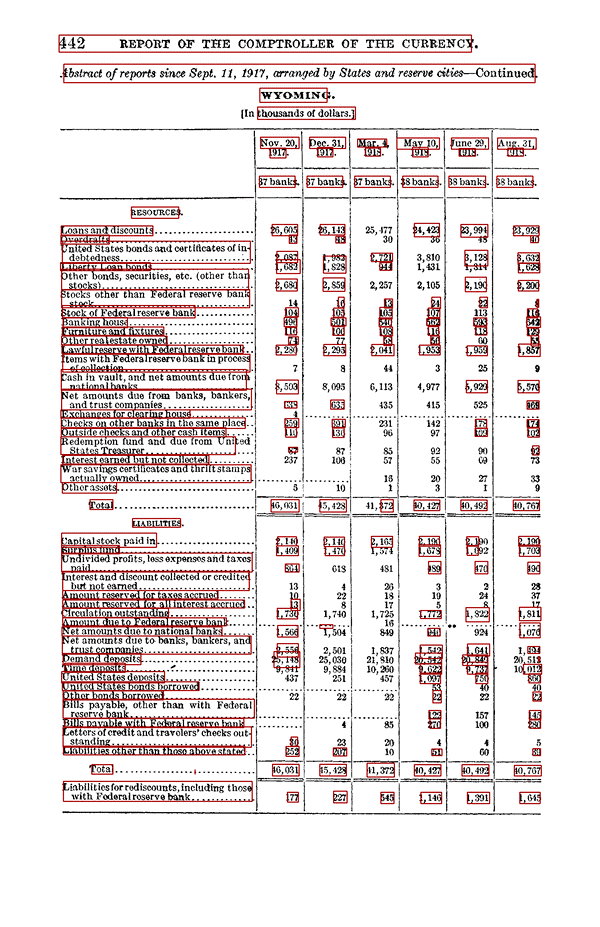}
    \caption{Combined lines (correct)}
    \label{fig:rows3}
  \end{subfigure}%\hspace*{-0.1em}
  \caption{\textbf{Identifying balance sheet labels. }To correctly identify rows---and thus balance sheet items---of each
table, we apply a three-part process. First, we identify the
\emph{words} recognized by an OCR engine such as Google Vision AI.
Second, we concatenate words based on their relative horizontal
distance. Lastly, we concatenate lines based on their indenting and
vertical distance and are thus able to correctly identify row labels.}
  \label{fig:table-rows}
\end{figure}
}

\hypertarget{data-invariants-and-sanity-checks}{%
\subsubsection{Data invariants and sanity
checks}\label{data-invariants-and-sanity-checks}}

Another important way of improving the digitization process is to flag
potential errors, which can be then reviewed and corrected by hand.
Without such flags, even a detailed human reviewer may overlook
potentially problematic mistakes in the data.

This is particularly important for panel balance sheet data, which has
two particular properties. First, errors are very costly in such
datasets. For instance, suppose a bank has stable total assets of \$100,
but in one year there is a typo that adds one zero at the end, so the
reported value of total assets becomes \$1000 for that year. This
mistake would lead to an incorrect 1000\% increase in total assets in
the year where it occurs, plus another incorrect value of the subsequent
year, where total assets would be reported to fall by 90\%. Second,
balance sheets have several constraints or invariants that we could
exploit to validate the quality of the data. For instance, the balance
sheet identity must hold, so the sum of total assets must equal the
``total assets'' label, which in turn must equal the ``total liabilities
and equity'' label.

To increase confidence in the quality of the digitized data and to help
guide human reviewers, the practitioner must identify as many
constraints as possible and incorporate them into the code. For
instance, the following are some of the constraints we implemented for
the OCC digitization project:

\begin{enumerate}
\def\labelenumi{\arabic{enumi}.}
\tightlist
\item
  Accounting identity. The sum of all assets must equal the label
  ``total assets''; the sum of all liability and equity labels must
  equal the label ``total liabilities and equity''; the two total fields
  must be equal to each other.
\item
  Reserve requirements, bond-holding requirements, and minimum-capital
  requirements. Whenever such a constraint is violated, the page can be
  automatically flagged for human review.
\item
  Constraints across time. It was difficult to change the amount of
  certified bank capital during the National Banking Era. Hence, a
  change across time for the same bank can be flagged and reviewed.
\item
  Label and numeric constraints. Balance sheet labels that are not part
  of the valid label list, as well as invalid numeric fields, are
  automatically flagged. For instance, the number ``0123'' is always
  flagged because dollar values do not have a leading zero digit.
\item
  Unique identifiers. Bank charter numbers should uniquely identify
  banks across time, so a manual review was prompted if they were either
  missing or duplicated in a given year.
\end{enumerate}

Altogether, these constraints allow us to flag the most prominent errors
in the dataset, which then in turn releases more time for a more careful
inspection of the documents afterward.

\hypertarget{human-review}{%
\subsection{Human review}\label{human-review}}

In large-scale digitization projects, the human review step needs
special consideration as this is often the most expensive and
time-consuming part. In general, user productivity increases with
well-designed interfaces, that are comprehensible, predictable, and
controllable \citep{Shneiderman}. For our goals, we have found that the
most important thing is to ensure human reviewers are \emph{on task} for
as much time as possible. Any interruptions of the workflow such as
having to save corrected files, load and scroll through PDFs, or
manually \passthrough{\lstinline!alt-tab!} to review the list of pages
pending review will slow down the reviewers beyond just the amount of
time used by those interruptions.

Instead, we find that automating the human review process as much as
possible increases the attention available to reviewers, leading to
faster reviews with fewer errors. In terms of software, for both the OCC
and \emph{Saling's} project, we rely on an Excel workbook containing VBA
functions automating steps previously done manually by users. We have
also used a browser-based solution in an ongoing project digitizing data
from St.~Louis Fed's FRASER archives. While the Excel approach worked
quite well for the authors, the browser solution worked better when
delegating review tasks to others, as it reduced setup costs as well as
the sharing of input pages and corresponding corrections.
\Cref{fig-review} shows screenshots of both of these tools.

% Figure generated by panflute filter "media.py"
{
\setstretch{1.0}
\begin{figure}[ht!]
  \centering
  %%%% Panel 1 %%%%
  \begin{subfigure}{0.6\textwidth}
    \centering
    \includegraphics[width=0.9\linewidth]{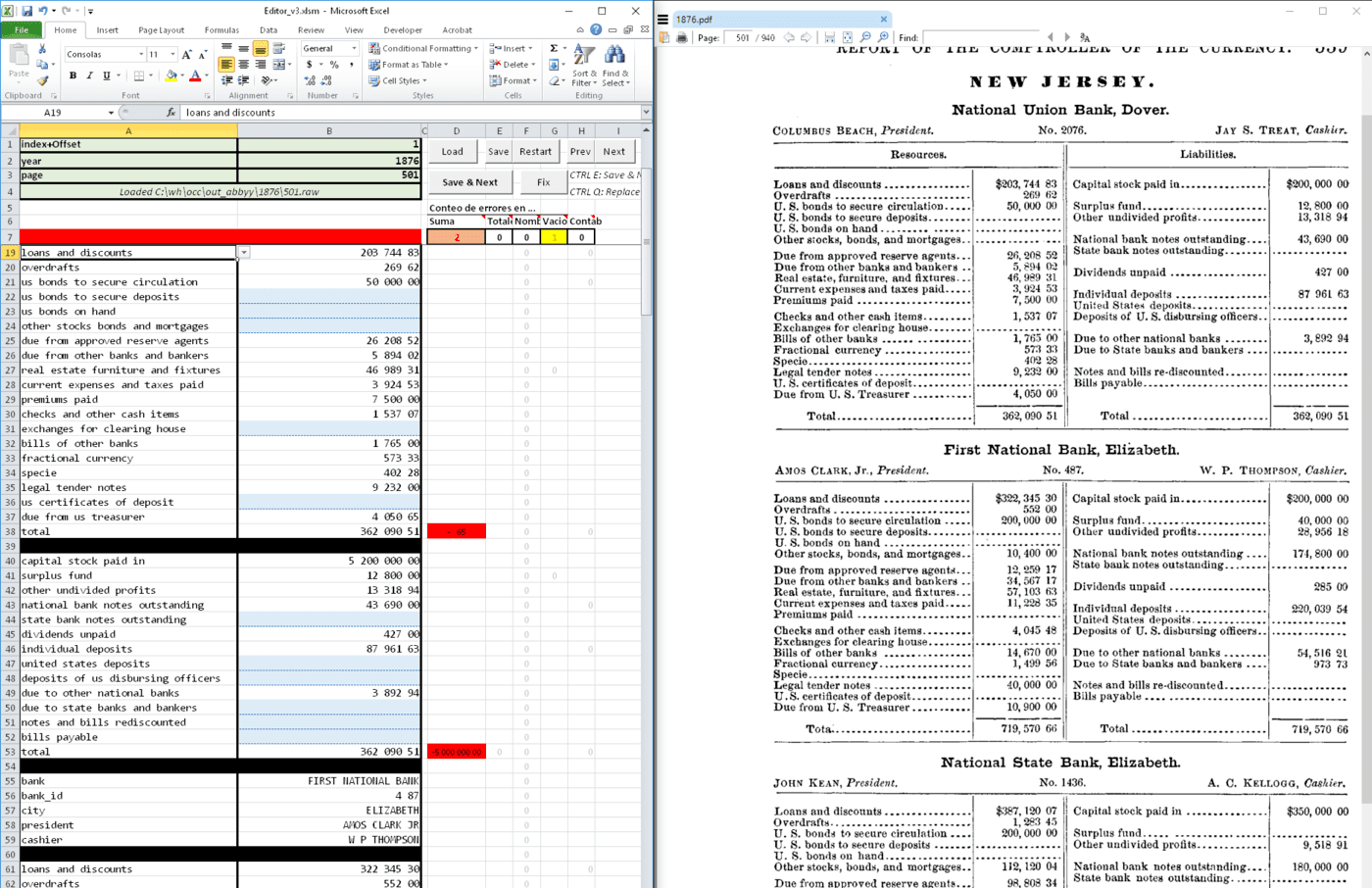}
    \caption{OCC Dataset}
    \label{occ-dataset}
  \end{subfigure}%\hspace*{-0.1em}
\hfill
  %%%% Panel 2 %%%%
  \begin{subfigure}{0.6\textwidth}
    \centering
    \includegraphics[width=0.9\linewidth]{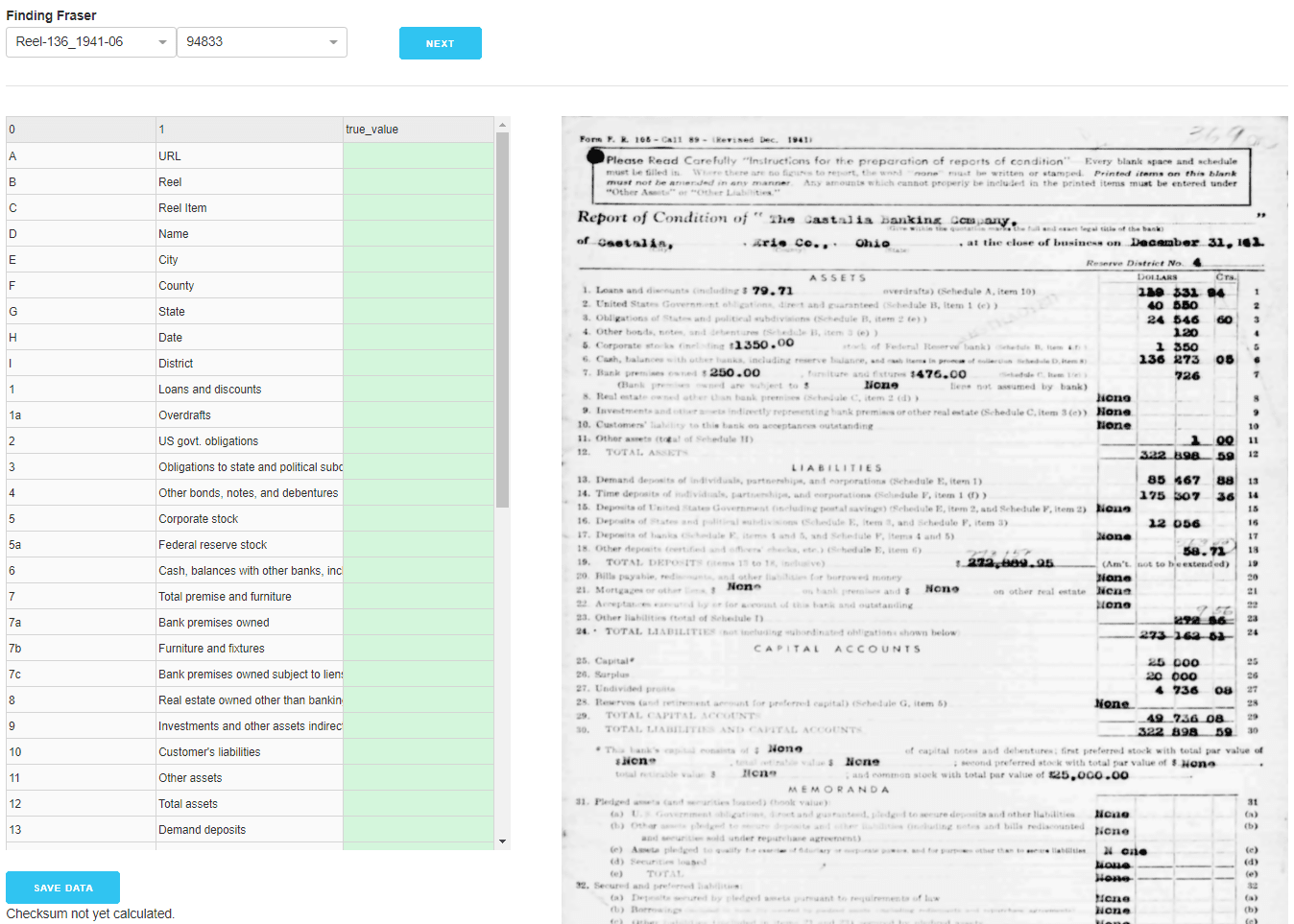}
    \caption{FRASER Dataset}
    \label{fraser-dataset}
  \end{subfigure}%\hspace*{-0.1em}
  \caption{\textbf{Interfaces for human review of transcription output. }Panel (a) shows a screenshot of the program used to validate the OCC and
\emph{Saling's} dataset. It consists of an Excel workbook powered by a
set of VBA functions that load and save the data, and display the
corresponding images. Panel (b) shows a screenshot of an ongoing program
used to validate St.~Louis Fed's FRASER archives. It is a self-contained
website powered by Python Dash.}
  \label{fig-review}
\end{figure}
}

Based on our experiences, below we list some of the steps where we found
automating was invaluable:

\begin{enumerate}
\def\labelenumi{\arabic{enumi}.}
\tightlist
\item
  Show data and scanned images side-by-side. Whenever the reviewer loads
  a new page or table, the corresponding image should be automatically
  loaded as well.
\item
  Add shortcuts as much as possible, for loading and saving CSV files,
  navigating the list of pages that need to be reviewed, etc.
\item
  Flagging likely errors in red or yellow colors, so the reviewer can
  monitor those items more closely.
\item
  Automatically keeping track of the balance sheet identity, so the
  reviewer can know when errors are remaining.
\end{enumerate}

Once we automated these steps, we found a substantial increase in review
speed, leading to a larger amount of ground-truth data, which in turn
can be utilized to improve the overall digitization pipeline, as
discussed below.

\hypertarget{parameter-tuning}{%
\subsubsection{Parameter tuning}\label{parameter-tuning}}

Parameter tuning---or more appropriately, hyper-parameter tuning, as it
is known in the machine learning community---is a method for selecting
the optimal values for the parameters employed throughout the
digitization process. It involves using ground truth data, known to be
correct, to create a measure of the overall quality of the digitization
process, and then ``tuning the parameters'' to maximize the value of
this measure.

At its simplest, this tuning could be done manually. That is what we do
in \cref{fig:tuning}, where we implemented a grid search to select the
optimal binarization threshold of the fore-edge removal step
(\cref{fig:fore-edges}).

% Figure generated by panflute filter "media.py"
{
\setstretch{1.0}
  \begin{figure}[htpb]
    \centering
    \includegraphics[width=0.8\textwidth]{"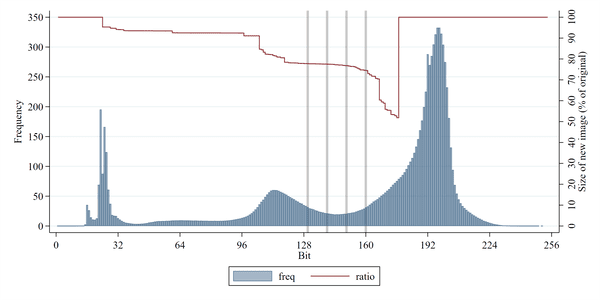"}
    \caption{\textbf{Fine-tuning pre-processing parameters. }This figure shows how the binarization parameter used in
\cref{fig:fore2} affects the performance of the trimming step shown in
\cref{fig:fore-edges}. The histogram in blue shows the distribution of
all points in a scanned page, from 0 (black) to 255 (white). The three
modes of the histogram correspond to the black background and fonts, the
dark gray book edge, and the light gray page background. The red line
shows how changing the binarization parameter affects how much of the
page is trimmed. Choosing a number between 128 and 160 will correctly
separate the page background from the book edges, so the edges can be
removed without trimming down the actual page contents.}
    \label{fig:tuning}
  \end{figure}
}

For more advanced uses, grid search can be replaced by more efficient
methods such as gradient descent. Moreover, if we expect to perform
large amounts of automatic parameter tuning, we also advise users to set
aside part of the ground truth data so it can be used for
cross-validation, which helps to prevent overfitting the algorithms to
set of ground truth data already collected.

\hypertarget{sec-code}{%
\section{Software implementation}\label{sec-code}}

This section describes the main functionality of the
\passthrough{\lstinline!quipucamayoc!} Python package, available online
at \url{https://github.com/sergiocorreia/quipucamayoc/}. The goal of
this package is to make the methods discussed in \cref{sec-internals}
accessible to researchers and lower the bar for future efforts to
digitize structured historical data.

\hypertarget{why-quipucamayoc}{%
\paragraph*{Why quipucamayoc}\label{why-quipucamayoc}}
\addcontentsline{toc}{paragraph}{Why quipucamayoc}

We have found that to make OCR algorithms truly accessible to
researchers and practitioners it is not enough to just provide a code
implementation of the algorithms themselves. There are significant
barriers to using these tools, such as difficult installation processes
and lack of compatibility with other OCR tools that researchers might be
using already. Thus, we built \passthrough{\lstinline!quipucamayoc!}
with two principles in mind:

\begin{enumerate}
\def\labelenumi{\arabic{enumi}.}
\tightlist
\item
  Batteries included. Providing a method that e.g., ``removes fore edges
  from a page'' is not particularly useful for a researcher that has no
  easy way of extracting images from PDFs or passing modified images to
  OCR engines for further processing. Thus, the first goal of this
  package is to provide all the tools required for a complete OCR
  pipeline, so users can get started easily.
\item
  Modular. For best performance on large datasets, users will likely
  want to extend the methods provided by
  \passthrough{\lstinline!quipucamayoc!} or integrate them with other
  algorithms available in the Python ecosystem. To facilitate this
  integration and extensibility, \passthrough{\lstinline!quipucamayoc!}
  is built in a modular way, so the different parts can be modified and
  extended independently of each other and connected with other machine
  learning or computer vision Python packages.
\end{enumerate}

\hypertarget{getting-started}{%
\paragraph*{Getting started}\label{getting-started}}
\addcontentsline{toc}{paragraph}{Getting started}

The package is available on the Python Package Index (PyPI), so it can
be installed on the command line via
\passthrough{\lstinline!pip install quipucamayoc!}. This step also loads
several other packages for numerical analysis, image processing, cloud
OCR integration, etc.

The main object in this package is the
\passthrough{\lstinline!Document!} class, which encapsulates all
pre-processing, OCR, and post-processing tools. It allows for two types
of inputs: PDF documents and folders with sequentially numbered images,
each representing a page in a scanned document. For instance, when
working with a PDF file, a common first step would be:

\begin{lstlisting}[language=Python, caption={Starting up}]
import quipucamayoc  # Import package
doc = quipucamayoc.Document('myfile.pdf', cache_folder='myfile-data')  # Validate that the file exists and is accessible
doc.describe()  # Display number of pages, metadata, etc.
doc.extract_images()  # Extract all pages of the PDF as images on the 'myfile-data' folder
\end{lstlisting}

Once the images are extracted, each image can be manipulated
individually:

\begin{lstlisting}[language=Python, caption={Working with pages}]
page = doc.pages[0]   # Select a page to process
page.load()           # Load image from disk into memory
page.view_image()     # Display current version of the image on screen
\end{lstlisting}

\hypertarget{code-image-processing}{%
\paragraph*{Image processing}\label{code-image-processing}}
\addcontentsline{toc}{paragraph}{Image processing}

This step is implemented as a set of methods that can be iteratively
applied to all pages, as shown in \cref{listing-image}. These methods
are, for the most part, based on the OpenCV and scikit-image libraries,
and are thus computationally efficient.

\begin{lstlisting}[language=Python, caption={Image processing}, label={listing-image}]
for page in doc.pages:
    page.load()
    page.remove_black_background()         # Crop black background
    page.remove_fore_edges(threshold=160)  # Remove book fore-edges
    page.dewarp()                          # Dewarp and deskew image
    page.convert('grayscale')              # Convert image to grayscale
    page.apply_clahe()                     # Improve contrast with CLAHE
    page.binarize(method='sauvola')        # Apply Sauvola's binarization
    page.save()                            # Save page into disk
\end{lstlisting}

\hypertarget{code-layout-recognition}{%
\paragraph*{Layout recognition}\label{code-layout-recognition}}
\addcontentsline{toc}{paragraph}{Layout recognition}

Similarly, layout recognition methods are also available at the page
level:

\begin{lstlisting}[language=Python, caption={Layout recognition}]
page.detect_lines(columns=5, save_annotated_image=True)
page.lines.view_annotated_image()  # View image with lines overlaid, for debugging
\end{lstlisting}

\hypertarget{ocr}{%
\paragraph*{OCR}\label{ocr}}
\addcontentsline{toc}{paragraph}{OCR}

For PDF documents with no pre-processing, the OCR step is available at
the document level, while for documents where images have already been
extracted and processed, OCR can be applied at the page level:

\begin{lstlisting}[language=Python, caption=OCR, label={listing-ocr}]
# Document OCR
doc.run_ocr(engine='google')

# Page OCR
page.run_ocr(engine='google')     # Synonyms: gcv, visionai
page.run_ocr(engine='amazon')     # Synonyms: aws, textract
page.run_ocr(engine='microsoft')  # Synonyms: azure (not currently implemented)
page.run_ocr(engine='tesseract')  # (not currently implemented)

# What OCR engines have been applied to a given page?
print(page.engines)
\end{lstlisting}

As shown in \cref{listing-ocr}, \passthrough{\lstinline!quipucamayoc!}
is designed so different OCR engines are as interchangeable as possible.

Further, some OCR engines are also able to extract additional elements
such as tables:

\begin{lstlisting}[language=Python, caption={Table extraction}, label={listing-textract}]
# Amazon's OCR engine involves a difficult initial setup
quipucamayoc.install_aws()

doc.extract_tables(engine='aws')
\end{lstlisting}

Because the target market for the OCR offerings of the commercial cloud
providers is certainly not researchers but instead large corporations
with dedicated IT \emph{divisions}, setting up these providers to
interact with \passthrough{\lstinline!quipucamayoc!} is a complex task.
For example,
\href{https://docs.aws.amazon.com/textract/latest/dg/api-async-roles.html}{substep
4 of step 6} of the instructions for Amazon's Textract requires
carefully giving ``IAM users or AWS accounts the appropriate permissions
to publish to the Amazon SNS topic and read messages from the Amazon SQS
queue.'' Because knowing the nuances of these steps is of no particular
value for economic historians, we have implemented functions that
programmatically link \passthrough{\lstinline!quipucamayoc!} with OCR
engines. For instance, the \passthrough{\lstinline!install\_aws()!}
function encapsulates all the steps required to setup Amazon's Textract
on a given AWS account and on a given
\passthrough{\lstinline!quipucamayoc!} installation.

\hypertarget{code-data-extraction-and-validation}{%
\paragraph*{Data extraction and
validation}\label{code-data-extraction-and-validation}}
\addcontentsline{toc}{paragraph}{Data extraction and validation}

This is the most customizable and user-dependent step, as it involves
rearranging the extracted characters or tables into a meaningful
dataset, which often requires expert knowledge of the researcher. For
instance, a user might write a \passthrough{\lstinline!process\_page()!}
function that takes as input a given page, with all its associated
information (recognized text, tables, etc.) and use it to output a CSV
file with validated information:

\begin{lstlisting}[language=Python, caption={Data extraction and validation}]
s = quipucamayoc.load_spellchecker(filename='frequencies.csv')  # Load a spellchecker based on a list of valid words
process_data(page, spellchecker=s)  # "process_data" is a user-created function

# Internals are accessible:
page.textract_tables[0][0][0]  # View cell A1 of the first table of the page
page.ocr['google'].paragraphs[0]  # View first paragraph of the page
\end{lstlisting}

\hypertarget{command-line-tools}{%
\paragraph*{Command-line tools}\label{command-line-tools}}
\addcontentsline{toc}{paragraph}{Command-line tools}

Some of the methods provided by \passthrough{\lstinline!quipucamayoc!}
are available as command-line tools. This allows practitioners to use
these methods and obtain OCR results without needing to write python
scripts. This makes this package more practical for small-scale
projects, where the researcher might want to run OCR to get to a
starting point from which she will correct and validate data by hand:

\begin{lstlisting}[language=bash, caption={Command-line usage}]
$ quipu install aws
(installing quipucamayoc pipeline for aws/textract)
$ quipu extract-tables myfile.pdf
(uploading file)
(waiting for OCR completion)
(downloading output)
(3 CSV files generated and placed on the ./myfile/ folder)
\end{lstlisting}

\hypertarget{future-development}{%
\paragraph*{Future development}\label{future-development}}
\addcontentsline{toc}{paragraph}{Future development}

\passthrough{\lstinline!quipucamayoc!} is an actively maintained and
developed package, and as such, is open to improvements from both the
authors as well as future users of the package. We believe potential
users will benefit from the existing tools and ``batteries included''
design; and for large-scale projects, we encourage them to extend the
set of pre- and post-processing tools to improve the quality of their
output.

\hypertarget{sec-summary}{%
\section{Summary}\label{sec-summary}}

Archives and libraries have vast amounts of structured historical data
of high value to economic historians, such as balance sheets, logs and
records, price lists, and so on. For the most part, these datasets are
considered either too expensive or too difficult to digitize for
researchers without access to large funds or arrays of research
assistants. In this paper, we propose that this is not the case anymore
and that even large-scale datasets can be accurately digitized with only
modest resources and limited programming knowledge. Although there is no
one-size-fits-all solution, we argue for leveraging well-established and
battle-tested tools, such as OpenCV and cloud-based OCR software, to
construct digitization pipelines that can be tailored to the data
sources at hand while requiring the least amount of customized ad-hoc
code.

For large-scale datasets, we suggest that researchers' time is more
valuable when applied to the elements specific to their datasets,
instead of focusing on directly managing the OCR tools and the
scaffolding connecting the different parts of the OCR pipeline. By
avoiding the more repetitive and cumbersome aspects of the OCR aspects,
researchers are more able to devote resources to developing metrics for
identifying errors in the data, either by constructing ground truths via
human reviews or by exploiting characteristics of the data at hand. In
the case of balance sheet records, we can exploit accounting identities
to allow for straightforward error detection. In contrast, security
price data may be less well suited, as they contain fewer constraints
that we can validate against. Moreover, accuracy metrics allow
researchers to easily test and tune the different components of the
digitization pipeline, so instead of being relegated to the latter
stages of the digitization, they can be used from the beginning to build
a more accurate pipeline.

Lastly, although the pipeline discussed in this paper can be quite
complex depending on the scale and difficulty of the digitization task,
we believe that for most use cases even simple pipelines would perform
quite accurately while maintaining only very minor programming
requirements. Indeed, we recommend starting with simpler pipelines even
for large projects, and only adding complexity as needed, which
maximizes the likelihood of successful digitization efforts and helps to
avoid premature optimization problems.

\hypertarget{refs}{}
\begin{CSLReferences}{0}{0}
\end{CSLReferences}

  \bibliography{literature}
\end{document}